\documentclass[journal]{IEEEtran}
\usepackage{cite}

\ifCLASSINFOpdf
\else
\fi
\usepackage{amsmath}
\usepackage{algorithmic}

\usepackage{array}

\ifCLASSOPTIONcompsoc
  \usepackage[caption=false,font=normalsize,labelfont=sf,textfont=sf]{subfig}
\else
  \usepackage[caption=false,font=footnotesize]{subfig}
\fi
\usepackage{dblfloatfix}

\ifCLASSOPTIONcaptionsoff
 \usepackage[nomarkers]{endfloat}
 \let\MYoriglatexcaption\caption
 \renewcommand{\caption}[2][\relax]{\MYoriglatexcaption[#2]{#2}}
\fi
\usepackage{url}

\usepackage{amssymb,amsfonts,mathtools}
\usepackage{amsthm,amscd,amsbsy}
\usepackage[ruled]{algorithm2e}

\usepackage{float}

\usepackage{graphicx,epsfig,color,graphics,caption}
\usepackage{epsf,psfrag,hhline,textcomp,enumerate}
\usepackage{xcolor}
\usepackage{times,fancyhdr}
\usepackage{multirow}
\usepackage{latexsym}
\usepackage{makecell}
\usepackage{comment}
\usepackage{tabularx} 
\usepackage{paralist}
\usepackage{booktabs}
\usepackage{hyperref}
\hypersetup{
	unicode=true,
	pdftitle={Explainable, Stable, and Scalable Graph Convolutional Networks for Learning Graph Representation},
	pdfauthor={Ping-En Lu, Cheng-Shang Chang},
	pdfsubject={IEEE, Transactions on Neural Networks and Learning Systems, IEEE Transactions on Neural Networks and Learning Systems, TNNLS, IEEE TNNLS},
	pdfcreator={Ping-En Lu},
	pdfkeywords={Graph convolutional networks, graph neural networks, network embedding, eigenvectors, GNN, GCN, GNNs, GCNs},
	colorlinks=true,
	linkcolor=blue
}

\usepackage[utf8]{inputenc}
\usepackage[english]{babel}
\usepackage[export]{adjustbox}
\usepackage{sidecap}
\usepackage{environ}
\newcommand{\mymarginpar}[1]{\marginpar{#1}}
\renewcommand{\marginpar}[1]{}

\newcommand{\bearn}{\begin{eqnarray*}}
	\newcommand{\eearn}{\end{eqnarray*}}
\newcommand{\barr}{\begin{array}}
	\newcommand{\earr}{\end{array}}

\newcommand \Q {{\bf Q}}
\newcommand \qq{q}
\newcommand \deltaone{\delta}

\newcommand{\N}{{\cal N}}

\newtheorem{definition}{Definition}
\newtheorem{assumption}[definition]{Assumption}
\newtheorem{property}[definition]{Property}
\newtheorem{proposition}[definition]{Proposition}
\newtheorem{lemma}[definition]{Lemma}
\newtheorem{theorem}[definition]{Theorem}
\newtheorem{corollary}[definition]{Corollary}
\newtheorem{example}[definition]{Example}
\newtheorem{remark}[definition]{Remark}
\newtheorem{axiom}[definition]{Axiom}

\newcommand{\benum}{\begin{enumerate}}
	\newcommand{\eenum}{\end{enumerate}}
\newcommand{\bdesc}{\begin{description}}
	\newcommand{\edesc}{\end{description}}

\newcommand{\bdefin}[1]{\begin{definition}\mymarginpar{def:#1}\label{def:#1}}
	\newcommand{\edefin}{\end{definition}}

\newcommand{\bpro}[1]{\begin{property}\mymarginpar{pro:#1}\label{pro:#1}}
	\newcommand{\epro}{\end{property}}

\newcommand{\bprop}[1]{\begin{proposition}\mymarginpar{prop:#1}\label{prop:#1}}
	\newcommand{\eprop}{\end{proposition}}

\newcommand{\blem}[1]{\begin{lemma}\mymarginpar{lem:#1}\label{lem:#1}}
	\newcommand{\elem}{\end{lemma}}

\newcommand{\bass}[1]{\begin{assumption}\mymarginpar{the:#1}\label{ass:#1}}
	\newcommand{\eass}{\end{assumption}}

\newcommand{\bthe}[1]{\begin{theorem}\mymarginpar{the:#1}\label{the:#1}}
	\newcommand{\ethe}{\end{theorem}}
\newcommand{\rthe}[1]{Theorem \ref{the:#1}}

\newcommand{\bcor}[1]{\begin{corollary}\mymarginpar{cor:#1}\label{cor:#1}}
	\newcommand{\ecor}{\end{corollary}}
\newcommand{\rcor}[1]{Corollary \ref{cor:#1}}
\newcommand{\bax}[1]{\begin{axiom}\mymarginpar{ax:#1}\label{ax:#1}}
	\newcommand{\eax}{\vspace{-.1in} \end{axiom}}

\newcommand{\bex}[2]{\vspace{.1in}\begin{example}\mymarginpar{ex:#1}{\bf #2}\label{ex:#1}}
	\newcommand{\eex}{\end{example}\vspace{.3cm}}

\newcommand{\brem}[1]{\begin{remark}\mymarginpar{rem:#1}\label{rem:#1}\em}
	\newcommand{\erem}{\end{remark}}

\newcommand{\beq}[1]{\mymarginpar{eq:#1}\begin{equation}\label{eq:#1}}
	\newcommand{\beqno}[1]{\mymarginpar{eq:#1}\begin{eqnarray}\nonumber}
		\newcommand{\eeq}{\end{equation}}
	\newcommand{\eeqno}{&&\end{eqnarray}}
\newcommand{\req}[1]{(\ref{eq:#1})}

\newcommand{\bear}[1]{\mymarginpar{eq:#1}\begin{eqnarray}\label{eq:#1}}
	\newcommand{\bearno}[1]{\mymarginpar{eq:#1}\begin{eqnarray}\nonumber}
		\newcommand{\eear}{\end{eqnarray}}
	\newcommand{\eearno}{\end{eqnarray}}
\NewEnviron{es}{
	\begin{equation}\begin{split}
			\BODY
	\end{split}\end{equation}
}
\newcommand{\bieeeeq}[1]{\mymarginpar{eq:#1}\begin{IEEEeqnarray}{rCl}\label{eq:#1}}
	\newcommand{\eieeeeq}{\end{IEEEeqnarray}}
\newcommand{\bsel}{\left \{\begin{array}{cl}}
	\newcommand{\esel}{\end{array}\right .}
\newcommand{\bmat}[1]{\left [\begin{array}{#1}}
	\newcommand{\emat}{\end{array}\right ]}

\newcommand{\rsec}[1]{Section \ref{sec:#1}}

\newcommand{\bapp}{\begin{appendices}}
	\newcommand{\eapp}{\end{appendices}}
\def\R{I\kern-0.30em R}
\def\N{I\kern-0.30em N}
\def\P{I\kern-0.30em P}

\newcommand{\rfig}[1]{Figure \ref{fig:#1}}

\usepackage{hhline}
\usepackage{makecell}
\newcolumntype{?}{!{\vrule width 1pt}}

\begin{document}
%
\title{Explainable, Stable, and Scalable Graph Convolutional Networks for Learning Graph Representation}
%
%
%

\author{Ping-En~Lu,~\IEEEmembership{Graduate~Student~Member,~IEEE,} and~Cheng-Shang~Chang,~\IEEEmembership{Fellow,~IEEE}
\thanks{The authors are with the Institute of Communications Engineering, National Tsing Hua University, Hsinchu 30013, R.O.C. (Taiwan).}
\thanks{E-mail: j94223@gmail.com, and cschang@ee.nthu.edu.tw.}
\thanks{This manuscript was submitted to IEEE Transactions on Neural Networks and Learning Systems (IEEE TNNLS) on September 22, 2020, and is being under reviewed. This work was supported by the Ministry of Science and Technology (MOST) of Taiwan (R.O.C.) under Project MOST108-2221-E007-016-MY3. (Corresponding author: Ping-En Lu.)}}

%
%

\markboth{This manuscript was submitted to IEEE Transactions on Neural Networks and Learning Systems on 09 22, 2020.}%
{Lu \MakeLowercase{\textit{et al.}}: Explainable, Stable, and Scalable Graph Convolutional Networks for Learning Graph Representation}
%



\maketitle

\begin{abstract}
The network embedding problem that maps nodes in a graph to vectors in Euclidean space can be very useful for addressing several important tasks on a graph. Recently, graph neural networks (GNNs) have been proposed for solving such a problem. However, most embedding algorithms and GNNs are difficult to interpret and do not scale well to handle millions of nodes. In this paper, we tackle the problem from a new perspective based on the equivalence of three constrained optimization problems: the network embedding problem, the trace maximization problem of the modularity matrix in a sampled graph, and the matrix factorization problem of the modularity matrix in a sampled graph. The optimal solutions to these three problems are the dominant eigenvectors of the modularity matrix. We proposed two algorithms that belong to a special class of graph convolutional networks (GCNs) for solving these problems: (i) Clustering As Feature Embedding GCN (CAFE-GCN) and (ii) sphere-GCN. Both algorithms are stable trace maximization algorithms, and they yield good approximations of dominant eigenvectors. Moreover, there are linear-time implementations for sparse graphs. In addition to solving the network embedding problem, both proposed GCNs are capable of performing dimensionality reduction. Various experiments are conducted to evaluate our proposed GCNs and show that our proposed GCNs outperform almost all the baseline methods. Moreover, CAFE-GCN could be benefited from the labeled data and have tremendous improvements in various performance metrics.
\end{abstract}

\begin{IEEEkeywords}
Graph convolutional networks (GCNs), graph neural networks (GNNs), network embedding, eigenvectors.
\end{IEEEkeywords}

%
\IEEEpeerreviewmaketitle

\section{Introduction}\label{labelofintroduction}
%
%
%
%
\IEEEPARstart{N}{etwork} embedding that learns a representation of a graph in a Euclidian space can be very useful for addressing several important problems, including link prediction, community detection (clustering), node classification, and graph classification. As such, it has attracted a lot of attention lately. In \cite{hamilton2017representation}, the authors provided a very good conceptual review of key advancements in this area of representation learning on graphs. These include matrix factorization-based methods (such as Laplacian Eigenmaps \cite{belkin2002laplacian}, Graph Factorization \cite{ahmed2013distributed}, GraRep \cite{cao2015grarep}, and HOPE \cite{ou2016asymmetric}), random-walk based algorithms (such as DeepWalk \cite{perozzi2014deepwalk} and node2vec \cite{grover2016node2vec}), and graph neural networks \cite{kipf2016semi, kipf2016variational}. However, as pointed out in the review paper \cite{hamilton2017representation}, there are still some open challenges. In particular, most of these network embedding algorithms do not scale well to handle millions of nodes. They are also difficult to interpret the physical meanings of the embedding vectors.

To tackle the scalability problem and the interpretability problem, in this paper, we propose the Clustering As Feature Embedding GCN (CAFE-GCN) algorithm that uses the clustering results to obtain embedding vectors. Our algorithm is based on two of our previous works: (i) the probabilistic framework of sampled graphs, and (ii) the softmax clustering algorithm.

\noindent {\bf The probabilistic framework of sampled graphs \cite{chang2015relative, chang2017probabilistic}}. Given a set of $n$ data points (called nodes in the paper) $\{u_1, u_2, \ldots, u_n\}$, the embedding problem is to map the $n$ data points to vectors in a Euclidean space so that points that are ``similar'' to each other are mapped to vectors that are close to each other. Such a problem is an ill-posed problem \cite{chang2019generalized} as people might have different views of ``similarity'' between two points (see, e.g., \cite{rosvall2008maps, lambiotte2014random, newman2018networks, hamilton2017representation}). In particular, Figure 1 of \cite{hamilton2017representation} provides a very insightful illustration for two different views of a character-character interaction graph derived from the Les Mis\'erables novel. To cope with such an ill-posed problem, one needs to specify a ``similarity'' measure at first. One commonly used ``similarity'' measure is a bivariate distribution $p_{U,W}(u,w)$ that measures the probability that the two points $u$ and $w$ are ``sampled'' together. A set of $n$ nodes $\{u_1, u_2, \ldots, u_n\}$ associated with a bivariate distribution $p_{U,W}(u,w)$ is called a {\em sampled graph} in \cite{chang2015relative, chang2017probabilistic}. The probabilistic framework of sampled graphs in \cite{chang2015relative, chang2017probabilistic} defines the notions of (relative) centrality, community, covariance, and modularity that can be interpreted intuitively. More details for the sampling methods, such as the uniform edge sampling, random walk sampling, and PageRank sampling \cite{PageRank}, can be found in \cite{chang2015relative, chang2017probabilistic}. One key result of this framework is that community detection (clustering) can be formulated as a modularity maximization problem.

\noindent {\bf The softmax clustering algorithm \cite{chang2017unified, chang2019generalized}}. Given a set of $n$ data points, the clustering problem is to cluster these data points so that data points within the same cluster are similar to each other and data points in different clusters are dissimilar. Thus, the embedding problem and the clustering problem are closely related \cite{belkin2002laplacian}. This is a well-known fact and a typical example is the spectral clustering algorithm (see, e.g., \cite{von2007tutorial}) that uses the eigendecomposition to embed data points into a Euclidean space and then uses the $K$-means clustering algorithm to cluster the embedded data points. There are also some previous works that use clustering algorithms for embedding data points (see, e.g., Softmax embedding \cite{chang2017unified}, GraphSAGE-GCN \cite{hamilton2017inductive}, DIFFPOOL\cite{ying2018hierarchical}, Generalized modularity embedding \cite{chang2019generalized} and LouvainNE \cite{bhowmick2020louvainne}). As in \cite{chang2017unified, chang2019generalized}, we use the softmax clustering algorithm to generate the embedding vectors. The key difference is that we take one step further to show that the softmax clustering algorithm is a special form of graph neural networks and {\em the embedding vectors} generated this way are actually approximations of dominant eigenvectors of the modularity matrix.

The development of the CAFE-GCN algorithm relies on the following four key insights:

\noindent {\bf (i) Equivalence of the embedding problem in a sampled graph.} We first formulate the embedding problem in a sampled graph as a constrained optimization problem and show that it is equivalent to a trace maximization problem and a matrix factorization problem. The matrix factorization formulation allows us to view the embedding problem in a sampled graph as an autoencoder and solve it by using matrix factorization methods. On the other hand, the trace maximization formulation allows us to view the embedding problem in a sampled graph as a modularity maximization problem and solve it by using well-known modularity maximization algorithms. More importantly, the trace maximization formulation shows that the optimal embedding can be found from the dominant eigenvectors of the modularity matrix (as a result of the Rayleigh-Ritz theorem).

\noindent {\bf (ii) Softmax clustering as a special form of GCN.} The softmax clustering algorithm in \cite{chang2017unified, chang2019generalized} is a special case of the Weisfeiler-Lehman (WL) algorithm \cite{weisfeiler1968reduction} that assigns the data points to colors (clusters) by using the softmax assignment. As such, it is also a special form of a more general class of graph neural networks \cite{scarselli2008graph, hamilton2017inductive}. Instead of using the (normalized) Laplacian matrix in the graph convolutional networks (GCNs)\cite{kipf2016semi}, CAFE-GCN uses the modularity matrix in the softmax clustering algorithm. The difference between using the Laplacian matrix and the modularity matrix is discussed in detail in Newman's book \cite{newman2018networks}. In particular, the Laplacian matrix is more suitable for graph cuts (with a known number of cuts) and the modularity matrix is more suitable for community detection where the number of communities is not known in advance. Moreover, the Laplacian matrix itself is not a ``similarity'' matrix \cite{wu2018tracking}. Its pseudo-inverse is. But it is very costly to compute the pseudo-inverse of a Laplacian matrix via the eigendecomposition.

\noindent {\bf (iii) Softmax clustering as a linear-time modularity maximization algorithm.} Let $m$ be the number of nonzero elements in the bivariate distribution $p_{U,W}(u,w)$ of a sampled graph. Then the computational complexity of the softmax clustering algorithm is $O(n+m)$ for each round of training $n$ nodes. If the sampled graph is sparse, i.e., $m=O(n)$, then the softmax clustering algorithm is a linear-time modularity maximization algorithm that converges monotonically to a local optimum.

\noindent {\bf (iv) Softmax embedding as approximations of dominant eigenvectors.} The dominant eigenvectors of the modularity matrix form the optimal embedding that maximizes the modularity. When $n$ is small, one can find the dominant eigenvectors by the power (orthogonal iteration) method (see, e.g., \cite{golub2013matrix}) with $O(n^3)$ computational complexity and $O(n^2)$ memory complexity \cite{hammond2011wavelets}. When the softmax clustering algorithm converges, it achieves a local maximum of the modularity that partitions the $n$ data points into clusters. If the local maximum is very close to the global optimum, then it is possible to find good approximations of the optimal embedding, i.e., the dominant eigenvectors, from the partition. The idea for this is to do one additional step of the orthogonal iteration.

In addition to CAFE-GCN, we propose another GCN, called sphere-GCN, that is also based on a modularity maximization algorithm. As such, the outputs of the sphere-GCN are also approximations of dominant eigenvectors. Like CAFE-GCN, there exists a linear-time implementation when the graph is sparse. The major difference between CAFE-GCN and sphere-GCN is that sphere-GCN embeds data points into a unit sphere, while CAFE-GCN embeds data points into probability vectors.

The contributions of this paper are summarized as follows:

\noindent {\bf (i)} To the best of our knowledge, CAFE-GCN and sphere-GCN are the first scalable, stable, and explainable GCN in the literature. They are linear-time algorithms that output approximations of dominant eigenvectors of the modularity matrix of a sampled graph.

\noindent {\bf (ii)} We conduct a theoretical analysis for CAFE-GCN and derive a theoretical bound between the approximation by CAFE-GCN and the largest eigenvector.

\noindent {\bf (iii)} We show that both CAFE-GCN and sphere-GCN can also be used to solve the dimensionality reduction problem and obtain approximations of the eigenvectors from PCA.

\noindent {\bf (iv)} We also propose a multi-layer CAFE-GCN algorithm that outputs multi-resolution embedding vectors for a sampled graph. The multi-layer CAFE-GCN algorithm is a nearly linear-time algorithm.

\noindent {\bf (v)} By conducting extensive numerical studies, we show that both CAFE-GCN and sphere-GCN are very effective in producing good approximations of dominant eigenvectors. Also, for various experimental settings on the three datasets, Cora \cite{sen2008collective}, Wiki \cite{yang2015network}, and ego-Facebook \cite{leskovec2012learning}, our proposed algorithms outperform almost all the baseline methods in the literature, including Graph Factorization \cite{ahmed2013distributed}, DeepWalk \cite{perozzi2014deepwalk}, node2vec \cite{grover2016node2vec}, LINE \cite{tang2015line}, HOPE \cite{ou2016asymmetric}, GraRep \cite{cao2015grarep}, and SDNE \cite{wang2016structural}. In particular, for the node classification task on the Cora dataset \cite{papers_with_code}, the semi-supervised CAFE-GCN achieves almost the same accuracy as the state-of-the-art method without using any side information.

\section{Equivalence of the embedding problem in sampled graphs}\label{embedding}
Given a set of $n$ data points (called nodes in the paper) $\{u_1, u_2, \ldots, u_n\}$, the embedding problem is to map the $n$ data points to vectors in a Euclidean space so that points that are ``similar'' to each other are mapped to vectors that are close to each other. Such a problem is an ill-posed problem and one needs to specify a ``similarity'' measure at first. One commonly used ``similarity'' measure is a bivariate distribution $p_{U,W}(u,w)$ that measures the probability that the two points $u$ and $w$ are ``sampled'' together. In this paper, we assume that the bivariate distribution is {\em symmetric}, i.e.,
\beq{sym1111}
p_{U,W}(u,w)=p_{U,W}(w,u).
\eeq
A set of $n$ nodes $\{u_1, u_2, \ldots, u_n\}$ associated with a bivariate distribution $p_{U,W}(u,w)$ is called a {\em sampled graph} in \cite{chang2015relative, chang2017probabilistic}.

\begin{definition}\label{def:covariance}{\bf (Covariance, Community, and Modularity \cite{chang2015relative, chang2017probabilistic}))}
	For a sampled graph, the covariance between two nodes $u$ and $w$ is defined as follows:
	\begin{equation}\label{eq:exp8888}
		\qq (u,w)=p_{U,W} (u,w)-p_{U}(u) p_{W}(w).
	\end{equation}
	Define the modularity matrix $\Q = (\qq (u,w))$ be the $n\times n$ matrix with its $(u,w)^{th}$ element being $\qq(u,w)$ (as a generalization of the original modularity matrix in \cite{newman2004fast}). Moreover, the covariance between two sets $S_1$ and $S_2$ is defined as follows:
	\begin{equation}\label{eq:exp9999}
		\qq (S_1,S_2)=\sum_{u \in S_1}\sum_{w \in S_2}\qq (u,w).
	\end{equation}
	Two sets $S_1$ and $S_2$ are said to be positively correlated if $\qq (S_1,S_2)\ge 0$. In particular, if a subset of nodes $S \subset V$ is positively correlated to itself, i.e., $\qq(S,S)\ge 0$, then it is called a {\em community} or a {\em cluster}. Let ${\cal P}=\{S_k,k=1,2, \ldots, K\}$, be a partition of $V$, i.e., $S_k \cap S_{k^\prime}$ is an empty set for $k \ne k^\prime$ and $\cup_{k=1}^K S_k=V$. The modularity $\Q({\cal P})$ with respect to the partition $S_k$, $k=1,2, \ldots, K$, is defined as
	\begin{equation}\label{eq:index1111}
		Q({\cal P})=\sum_{k=1}^K \qq(S_k,S_k).
	\end{equation}
\end{definition}

\begin{figure}[!htbp]%
	\centering
	\includegraphics[width=1.0\columnwidth]{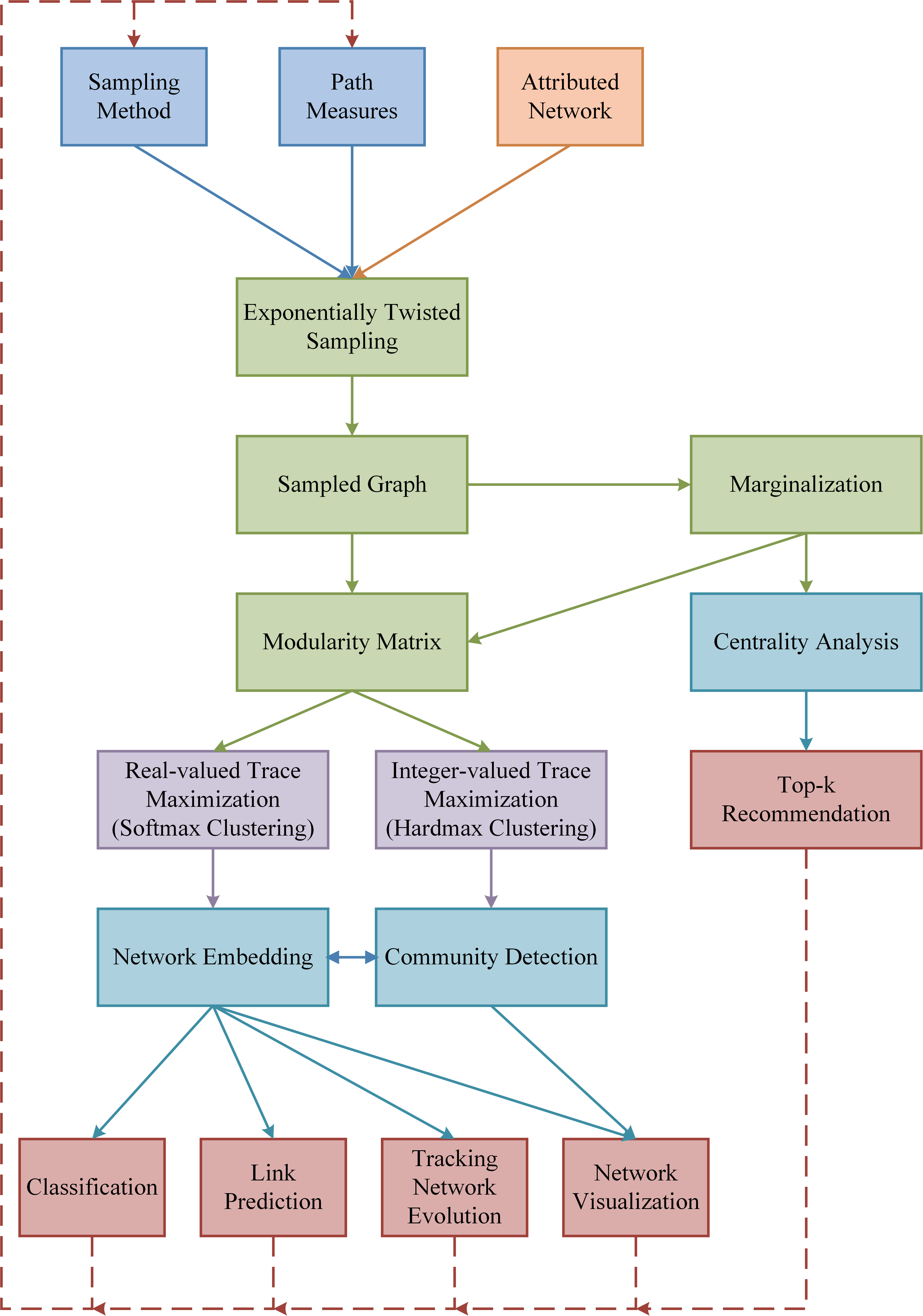}
	\caption{The probabilistic framework for centrality analysis, community detection, and network embedding in attributed networks.}
	\label{fig:flow}
\end{figure}

The probabilistic framework was extended to attributed networks in \cite{chang2018exponentially}, where nodes and edges can have attributes. The idea in \cite{chang2018exponentially} is to use exponentially twisted sampling along with path measures that incorporate the information of attributes into sampling. Once sampling is done and a bivariate distribution is obtained, one can then perform centrality analysis, community detection, and network embedding in attributed networks. In \rfig{flow}, we show the dependency graph (flow chart) of various tasks for analyzing attributed networks based on the probabilistic framework. Based on the framework, many tasks, including classification, link prediction, tracking network evolution, network visualization, and top-$k$ recommendation, can be developed and written in codes by using sampled graphs as inputs. In particular, community detection (clustering) can be formulated as a modularity maximization problem in \cite{chang2017probabilistic}. In this paper, we will show that the embedding problem can also be formulated as a modularity maximization problem.

We note that a bivariate distribution can be viewed as a normalized similarity measure. As pointed out in \cite{chang2015relative}, if there is a bounded similarity measure $sim(u,w)$ that gives a high score for a pair of two ``similar'' nodes $u$ and $w$, then one can map that similarity measure to a bivariate distribution $p_{U,W}(u,w)$ as follows:
\beq{link1111}
p_{U,W}(u,w)=\frac{sim(u,w)-\mbox{MINsim}}{\sum_{i=1}^n\sum_{j=1}^n\Big ( sim(i,j)-\mbox{MINsim} \Big )},
\eeq
where
\beq{link2222}
\mbox{MINsim}=\min_{1 \le i,j \le n}sim(i,j),
\eeq
is the minimum value of all the similarity scores. The advantage of using bivariate distributions is that we can have probabilistic insights on network analysis. A very interesting recent work on the embedding problem for a bipartite network \cite{huang2019universal} also used a bivariate distribution to characterize a user-item network. There they showed that the optimal embedding vectors can be interpreted as conditional expectations.

In order to map data points that are similar to each other to vectors that are close to each other, the embedding problem can be formulated as the optimization problem that minimizes the following weighted distance:
\beq{embed1111}
\sum_{u=1}^n \sum_{w=1}^n \qq(u,w) ||h_u-h_w||^2,
\eeq
where $h_u=(h_{u,1}, h_{u,2}, \ldots, h_{u,K})^T$ is the vector mapped by node $u$ in $\mathbb{R}^K$, and $||h_u-h_w||^2$ is the squared Euclidean distance between $u$ and $w$. To understand the intuition of the minimization problem in \req{embed1111}, note that $-1 \le \qq(u,w) \le 1$. Two nodes with a positive (resp. negative) covariance should be mapped to two vectors with a small (resp. large) distance. The embedding vector $h_u=(h_{u,1}, h_{u,2}, \ldots, h_{u,K})^T$ can be viewed as the ``feature'' vector of node $u$ and $h_{u,k}$ is its $k^{th}$ feature. In practice, it is preferable to have uncorrelated features. For this, we add the constraints
\beq{embed1122}
\sum_{u=1}^n h_{u,k_1} h_{u,k_2} =0,
\eeq
for all $k_1 \ne k_2$. Also, to have bounded values for these features, we also add the constraints
\beq{embed1133}
\sum_{u=1}^n h_{u,k} h_{u,k} =1,
\eeq
for all $k$.

For such an embedding problem, the following equivalent statements were shown in the book chapter \cite{chang2019generalized}.

\bthe{embed} (\cite{chang2019generalized}, Theorem 3)
Let $\Q =(\qq_{u,w})$ be an $n \times n$ symmetric matrix with all its row sums and column sums being $0$, and $H$ be the $n \times K$ matrix with its $u^{th}$ row being $h_u$.
\begin{description}
	\item[(i)] The embedding problem in \req{embed1111} with the constraints in \req{embed1122} and \req{embed1133} is equivalent to the following optimization problem:
	\bieeeeq{opti1111tr}
	&\max &\mbox{tr}(H^T \Q H)\\
	&s.t. &H^T H={\bf I}_K,
	\eieeeeq
	where ${\bf I}_K$ is the $K \times K$ identify matrix.
	\item[(ii)] The embedding problem in \req{embed1111} with the constraints in \req{embed1122} and \req{embed1133} is equivalent to the following optimization problem:
	\bieeeeq{opti1111decom}
	&\min &||\Q- H H^T||_2^2\\
	&s.t. &H^T H={\bf I}_K,
	\eieeeeq
	where $||A||_2$ is the Frobenius norm of the matrix $A$.
\end{description}
\ethe

From \rthe{embed}(i), we know that solving the embedding problem is equivalent to solving the trace maximization problem in \req{opti1111tr}. As stated in \cite{von2007tutorial, kokiopoulou2011trace}, a version of the Rayleigh-Ritz theorem shows that the solution of the trace maximization problem in \req{opti1111tr} can be found by solving the dominant eigenvectors of the matrix $\Q$. This is stated in the following corollary.

\bcor{embedcor}
For the embedding problem in \req{embed1111} with the constraints in \req{embed1122} and \req{embed1133}, let $\lambda_1, \lambda_2, \ldots, \lambda_K$ be the $K$ largest eigenvalues of the matrix $\Q$ and $v_k=(v_{k, 1}, v_{k, 2}, \ldots, v_{k, n})^T$ be the eigenvector of $\Q$ corresponding to the eigenvalue $\lambda_k$. Then $h_u=( v_{1,u}, v_{2,u}, \ldots, v_{K,u})^T$, $u=1,2, \ldots, n$, are the optimal embedding vectors.
\ecor

One interpretation of the matrix factorization problem in \rthe{embed} is the autoencoder interpretation in \cite{hamilton2017representation}. One encodes each row of the modularity matrix $\Q$ into the corresponding row of the matrix $H$ and then uses the inner product to decode (and reconstruct an approximation) for the matrix $\Q$. The error term $||\Q- H H^T||_2^2$ is known as the {\em loss} function.

\section{The CAFE-GCN algorithm}\label{softmax}
\subsection{The softmax clustering algorithm}\label{algorithm}
In \cite{chang2019generalized}, a softmax clustering algorithm was proposed for clustering a sampled graph with a symmetric modularity matrix (see Algorithm \ref{alg:Softmax}). The softmax clustering algorithm used the softmax function \cite{bishop2006pattern} to map a $K$-dimensional vector of arbitrary real values to a $K$-dimensional probability vector. The algorithm starts from a non-uniform probability mass function for the assignment of each node to the $K$ clusters. Specifically, let $h_{u,k}$ denote the probability that node $u$ is in cluster $k$. Then one repeatedly feed each node to the algorithm to learn the probabilities $h_{u,k}'s$. When node $u$ is presented to the algorithm, its expected covariance $z_{u,k}$ to cluster $k$ is computed for $k=1,2,\ldots, K$. Instead of assigning node $i$ to the cluster with the largest positive covariance (the simple maximum assignment in the literature), Algorithm \ref{alg:Softmax} uses a softmax function to update $h_{u,k}'s$. Such a softmax update increases (resp. decreases) the confidence of the assignment of node $u$ to clusters with positive (resp. negative) covariances. The ``training'' process is repeated until the objective value $\mbox{tr}(H^T \Q H)$ converges to a local optimum. The algorithm then outputs the corresponding soft assignment vector for each node. It is worth mentioning that Algorithm \ref{alg:Softmax} can also be used as a semi-supervised learning algorithm. In the case that the clusters (labels) of a certain subset of nodes are known in advance, these nodes will not be affected by the other nodes, and they can be assigned to the corresponding clusters at the beginning of the training process and stay there through the whole training process.
\begin{algorithm}[!htbp]%
	\KwIn{A symmetric modularity matrix $\Q=(q(u,w))$, the number of clusters $K$, and the inverse temperature $\theta>0$.}
	\KwOut{A $n \times K$ soft assignment (probability) matrix $H=(h_{u,k})$ for $n$ nodes.}

	\noindent {\bf (1)} Set $q(u,u)=0$ for all $u$.

	\noindent {\bf (2)} Initially, each node $u$ is assigned with a (non-uniform) probability mass function $h_{u,k}$, $k=1, 2, \ldots, K$ that denotes the probability for node $u$ to be in cluster $k$.

	\noindent {\bf (3)} For $u=1, 2, \ldots, n$

	\noindent {\bf (4)} For $k=1, 2, \ldots, K$

	\noindent {\bf (5)} Compute $z_{u,k}=\sum_{w\neq u}q(w,u) h_{w,k}$.

	\noindent {\bf (6)} Let $\tilde h_{u,k}=e^{\theta z_{u,k}}h_{u,k}$, and $c=\frac{1}{\sum_{\ell=1}^K \tilde h_{u,\ell}}$.

	\noindent {\bf (7)} Update $h_{u,k} \Leftarrow c \cdot {\tilde h_{u,k}}$.

	\noindent {\bf (8)} Repeat from Step (3) until there is no further change.

	\caption{The Softmax Clustering Algorithm}
	\label{alg:Softmax}
\end{algorithm}

If $H$ is a (hard) assignment matrix, then $H$ corresponds to a partition of the $n$ nodes and $\mbox{tr}(H^T \Q H)$ is the modularity of that partition. Since $H$ is only a soft assignment matrix (with each row being a probability vector), $\mbox{tr}(H^T \Q H)$ can be viewed as the expected modularity.

The softmax clustering algorithm is in fact a modularity maximization algorithm that increases the expected modularity after each update of the soft assignment matrix. This is stated in the following theorem.

\bthe{ObjIncreasing}(\cite{chang2019generalized}, Theorem 5)
Given a symmetric matrix $\Q=(q(u,w))$ with $q(u,u)=0$ for all $u$, the following objective value
\beq{objc1111}
\mbox{tr}(H^T \Q H)= \sum_{k=1}^K\sum_{u=1}^n\sum_{w=1}^n q(u,w)h_{u,k}h_{w,k}
\eeq
is increasing after each update in Algorithm \ref{alg:Softmax}. Thus, the objective values converge monotonically to a finite constant.
\ethe

\subsection{Softmax clustering as a special form of GCN}\label{special}
As mentioned in the previous section, the embedding problem can be solved by finding the dominant eigenvectors of the matrix $\Q$. When $n$ is small, this can be done by the power (orthogonal iteration) method (see, e.g., \cite{golub2013matrix}) with $O(n^3)$ computational complexity and $O(n^2)$ memory complexity \cite{hammond2011wavelets}. In \cite{hammond2011wavelets}, a fast Chebyshev polynomial approximation algorithm was proposed to avoid the need for eigendecomposition of the matrix $\Q$, and that motivated Kipf and Welling \cite{kipf2016semi} to propose Graph Convolutional Networks (GCN) for semi-supervised classification. A GCN obtains the embedding vectors by carrying out the following iterations:
\beq{softmax1111}
H^{(\ell+1)}=\sigma(\Q H^{(\ell)}W^{(\ell)}),
\eeq
where $W^{(\ell)}$'s are trainable weight matrices and $\sigma$ is an activation function used in a neural network. As such, GCN can be viewed as a special class of graph neural networks (GNNs) in \cite{scarselli2008graph}. In this paper, we do not need the trainable weight matrices $W^{(\ell)}$'s and they are removed from \req{softmax1111} (or treated as the identity matrix). This leads to the following simplified GCN:
\beq{softmax1111b}
H^{(\ell+1)}=\sigma(\Q H^{(\ell)}).
\eeq
Instead of using the ReLU function in \cite{kipf2016semi}, we use the softmax function as our activation function. The softmax function $\sigma$ with $K$ inputs $z_1, z_2, \ldots, z_K$ generate the $K$ outputs
\beq{softmax2222}
\sigma(z_1, z_2, \ldots, z_K)=\frac{1}{\sum_{\ell=1}^K e^{\theta z_\ell}}(e^{\theta z_1}, e^{\theta z_2}, \ldots, e^{\theta z_K}),
\eeq
where $\theta>0$ is the inverse temperature. One nice feature of using the softmax function is that now every row of $H^{(\ell)}$ is a probability vector (with all its $K$ nonnegative elements summing to $1$). This leads to a probabilistic explanation of how the GCN in \req{softmax1111b} works. Let
\beq{softmax2255}
h_u^{(\ell)}=(h_{u,1}^{(\ell)},h_{u,2}^{(\ell)}, \ldots, h_{u,K}^{(\ell)})^T
\eeq
be the transpose of the $u^{th}$ row of the matrix $H^{(\ell)}$. As pointed out in \cite{kipf2016semi}, one can view the GCN as a special case of the Weisfeiler-Lehman (WL) algorithm \cite{weisfeiler1968reduction} that assigns the $n$ nodes to $K$ colors (or clusters). The probability $h_{u,k}^{(\ell)}$ then represents the probability that the $u^{th}$ node is assigned with color $k$. The GCN starts from a non-uniform probability mass function for the assignment of each node to the $K$ colors. Then we repeatedly feed each node to the GCN to learn the probabilities $h_{u,k}^{(\ell)}$'s. When node $u$ is presented to the GCN, its expected covariance
\beq{softmax3333}
z_{u,k}^{(\ell)}=\sum_{w=1}^n \qq(u,w)h_{u,k}^{(\ell)}
\eeq
to color $k$ is computed for $k=1,2,\ldots, K$. Instead of assigning node $u$ to the color with the largest positive covariance (the simple maximum assignment), GCN uses a softmax function to update $h_{u,k}^{(\ell)}$'s. Such a softmax update increases (resp. decreases) the confidence of the assignment of node $u$ to colors with positive (resp. negative) covariances. The ``training'' process is repeated until it converges.

A sequential implementation of the GCN in \req{softmax1111b} (like the usual training process of a neural network) is exactly the same as the softmax clustering algorithm in Algorithm \ref{alg:Softmax}.

\subsection{Softmax clustering as a linear-time modularity maximization algorithm}\label{linear}
Let $m$ be the number of nonzero elements in the bivariate distribution $p_{U,W}(u,w)$ of a sampled graph. We say a sampled graph is sparse if $m=O(n)$. Though Algorithm \ref{alg:Softmax} appears to be a matrix-based method, there is a linear-time implementation for a sparse sampled graph by using the techniques outlined in Section IV.B and IV.C of \cite{chang2017probabilistic}. To see this, note that the modularity matrix $\Q$ (though not sparse) can be decomposed as the difference of a sparse matrix $P=(p_{U,W}(u,w))$ and a rank $1$ matrix. Since the updates in \req{softmax3333} are only made {\em locally} (with $p_{U,W}(u,w) >0$), the computational complexity of the softmax clustering algorithm is $O(n+m)$ for each round of training $n$ nodes. As such, for a sparse sampled graph, the softmax clustering algorithm is a linear-time modularity maximization algorithm that converges monotonically to a local optimum.

\subsection{Softmax embedding as approximations of dominant eigenvectors}\label{approx}
In view of \rthe{ObjIncreasing}, the sequential implementation of the GCN in \req{softmax1111b} outputs a soft assignment matrix $H$ that achieves a local optimum of $\mbox{tr}(H^T \Q H)$. If we carry out one step of the orthogonal iteration, then we should be able to obtain an orthogonal matrix $\hat H$ that is closer to the optimal embedding for the trace maximization problem in \rthe{embed}(i). Specifically, we compute the QR decomposition for the $n \times K$ matrix $\Q H$ to find an $n \times K$ matrix ${\hat H}=({\hat h}_{u,k})$ and an $n \times n$ upper triangular matrix $R$ so that ${\hat H}R=\Q H$. Now the matrix $\hat H$ is an orthogonal matrix that satisfies the constraint ${\hat H}^T {\hat H}={\bf I}_K$. As shown in \rcor{embedcor}, the $K$ columns of the optimal embedding matrix are the $K$ eigenvectors corresponding to the largest $K$ eigenvalues. The matrix $\hat H$ is then an approximation of the $K$ eigenvectors corresponding to the largest $K$ eigenvalues. The detailed steps for obtaining the matrix $\hat H$ are shown in the CAFE-GCN algorithm in Algorithm \ref{alg:CAFE}.

Similar to Algorithm \ref{alg:Softmax}, Algorithm \ref{alg:CAFE} can also be used as a semi-supervised learning algorithm, denoted by CAFE-GCN (semi-supervised) in our experiments in \rsec{experi}, when there is a subset of known labels. In the extreme case that all the labels are known, we even do not need to perform the softmax clustering algorithm in Steps (1) and (2). As such, we can go directly to the QR decomposition in Step (3) of Algorithm \ref{alg:CAFE}. That leads to a speedy method of obtaining the embedding vectors for a dataset with known labels. Such an algorithm is denoted by CAFE-GCN (full label) in our experiments in \rsec{experi}.
\begin{algorithm}[!htbp]%
	\KwIn{A symmetric modularity matrix $\Q=(q(u,w))$, the maximum number of clusters $K$, and the inverse temperature $\theta>0$.}

	\KwOut{The number of clusters $C$, an $n \times C$ soft assignment matrix $H$, and an $n \times C$ embedding matrix $\hat H$ of nodes with $C \le K$.}

	\noindent {\bf (1)} Run the softmax clustering algorithm in Algorithm \ref{alg:Softmax} with the modularity matrix $\Q=(q(u,w))$, the number of clusters $K$, and the inverse temperature $\theta$.

	\noindent {\bf (2)} Let $C$ be the number of nonzero columns in $H$. Remove the zero columns in $H$. This yields a new $n \times C$ matrix $H$.

	\noindent {\bf (3)} Compute the QR decomposition for the $n \times C$ matrix $\Q H$ to find an $n \times C$ matrix ${\hat H}=({\hat h}_{u,k})$ and an $n \times n$ upper triangular matrix $R$ so that ${\hat H}R=\Q H$.

	\noindent {\bf (4)} Output the number of clusters $C$, the assignment matrix $H$, and the embedding matrix $\hat H$.
	
	\caption{The CAFE-GCN Algorithm}
	\label{alg:CAFE}
\end{algorithm}

\subsection{Theoretical bounds and numerical results for CAFE-GCN with $\bf K=2$}\label{results}
In this section, we conduct a theoretical analysis for CAFE-GCN with $K=2$ and derive a theoretical bound between the approximation by CAFE-GCN and the largest eigenvector.

For $K=2$, we have $h_{u,1}+h_{u,2}=1$ for all $u$ (as each embedding vector is a probability vector). Using this in \req{objc1111} yields
\bieeeeq{objc1111b}
&& \mbox{tr}(H^T \Q H)\nonumber\\
&&= \sum_{u=1}^n\sum_{w=1}^n q(u,w)h_{u,1}h_{w,1}+\sum_{u=1}^n\sum_{w=1}^n q(u,w)h_{u,2}h_{w,2}\nonumber\\
&&=2 \sum_{u=1}^n\sum_{w=1}^n q(u,w)h_{u,1}h_{w,1},
\eieeeeq
where we use the assumption that all the row sums and column sums of the matrix $\Q$ are equal to $0$. As a result of \rthe{ObjIncreasing} and \rthe{embed}, the CAFE-GCN with $K=2$ obtains a local optimum solution for the one-dimensional trace maximization in \req{opti1111tr}. This implies that
$$\sum_{u=1}^n\sum_{w=1}^n q(u,w)\frac{h_{u,1}}{\sqrt{\sum_{\ell=1}^n h_{\ell,1}^2}}\frac{h_{w,1}}{\sqrt{\sum_{\ell=1}^n h_{\ell,1}^2}}$$
should be very close to the largest eigenvalue of the matrix $\Q$. This motivates us to consider the $n$-vector $x=(x_1, x_2,\ldots x_n)^T$, where
\beq{embed6666}
x_u=\frac{h_{u,1}}{\sqrt{\sum_{\ell=1}^n h_{\ell,1}^2}}.
\eeq

Now we show that $x$ in \req{embed6666} is close to the eigenvector corresponding to the largest eigenvalue of $\Q$.
Recall that $\Q=(\qq(u,w))$ is a symmetric matrix with all its row sums and column sums being $0$. It is well-known that a real symmetric matrix is diagonalizable by an orthogonal matrix. Specifically, let $$\lambda_1 > \lambda_2 \ge \lambda_3 \ge \ldots \ge \lambda_n$$ be the (ordered) $n$ eigenvalues of $\Q$ and $v_i$ be the normalized (column) eigenvector corresponding to the eigenvalue $\lambda_i$, $i =1,2,,\ldots, n$. Let $V=[v_1 v_2 \ldots v_n]$ be the $n \times n$ orthogonal matrix formed by grouping the $n$ eigenvectors together. Then
\beq{diag1111}
V^T \Q V=D,
\eeq
where $D$ is diagonal matrix with the $n$ diagonal elements, $\lambda_1, \lambda_2, \ldots, \lambda_n$.

For our analysis, we also assume that there is a spectral gap between the largest eigenvalue and the second-largest eigenvalue magnitude (SLEM), i.e., ${\lambda_1}>\max[\lambda_2, -\lambda_n]$. Since $\Q$ has an eigenvalue $0$ with the eigenvector $\bf e$ (with all its elements being $1$), we know from the spectral gap assumption that $\lambda_1 > \lambda_2 \ge 0$. Moreover, as two eigenvectors corresponding two different eigenvalues are orthogonal for a real symmetric matrix, we then have $v_1^T {\bf e}=0$.

To measure the difference between two vectors, we use the {\em cosine similarity} defined below.

\bdefin{cosine}
The cosine similarity between two $n$-vectors $y$ and $z$, denoted by $COS(y,z)$, is
\beq{cos1111}
COS(y,z)=\frac{y^T z}{\sqrt{{y}^T y}\sqrt{z^T z}}.
\eeq
\edefin

In particular, if both $y$ and $z$ are unit vectors, i.e., $y^T y=z^T z=1$, then the Euclidean distance between these two vectors is $$\sqrt{(y-z)^T (y-z)}=\sqrt{2(1-COS(y, z))}.$$ As such, if the cosine similarity between two unit vectors is close to $1$, then the Euclidean distance between these two unit vectors is close to $0$.

In the following theorem, we show a lower bound for the cosine similarity between the vector $x$ in \req{embed6666} and $v_1$. Its proof is given in Appendix \ref{p_max} of this paper.

\bthe{max}
Let $\deltaone$ the ratio of the SLEM to the largest eigenvalue of $\Q$, i.e.,
\beq{gap1111}
\deltaone=\frac{\max[\lambda_2, -\lambda_n]}{\lambda_1}.
\eeq
Consider the vector $x$ in \req{embed6666}. If
\beq{max1100}
{x^T \Q x} \ge \lambda_1(1-\epsilon)
\eeq
for some $\epsilon$ satisfying
\beq{max0055}
0 \le \epsilon \le 1-\deltaone,
\eeq
then
\beq{max1111}
COS(v_1, x) \ge \sqrt{\frac{1-\epsilon-\deltaone}{1-\deltaone}}.
\eeq
Moreover, $COS(v_1, \Q x) \ge COS(v_1, x)$ and
\beq{max5555}
COS(v_1, \Q x)\ge \sqrt{\frac{1-\epsilon-\deltaone}{1-\epsilon-\deltaone+\deltaone^2}}.
\eeq
\ethe

\rthe{max} shows that if the CAFE-GCN obtains a good solution for the trace maximization problem in the sense of \req{max1100} and \req{max0055}, then it is close to the dominant eigenvector $v_1$ in terms of the bound of the cosine similarity in \req{max1111}. Moreover, the vector $\Q x$ is even closer to $v_1$, and it is even a better approximation of $v_1$.

\begin{figure}[!htbp]%
	\centering
	\includegraphics[width=1.0\columnwidth]{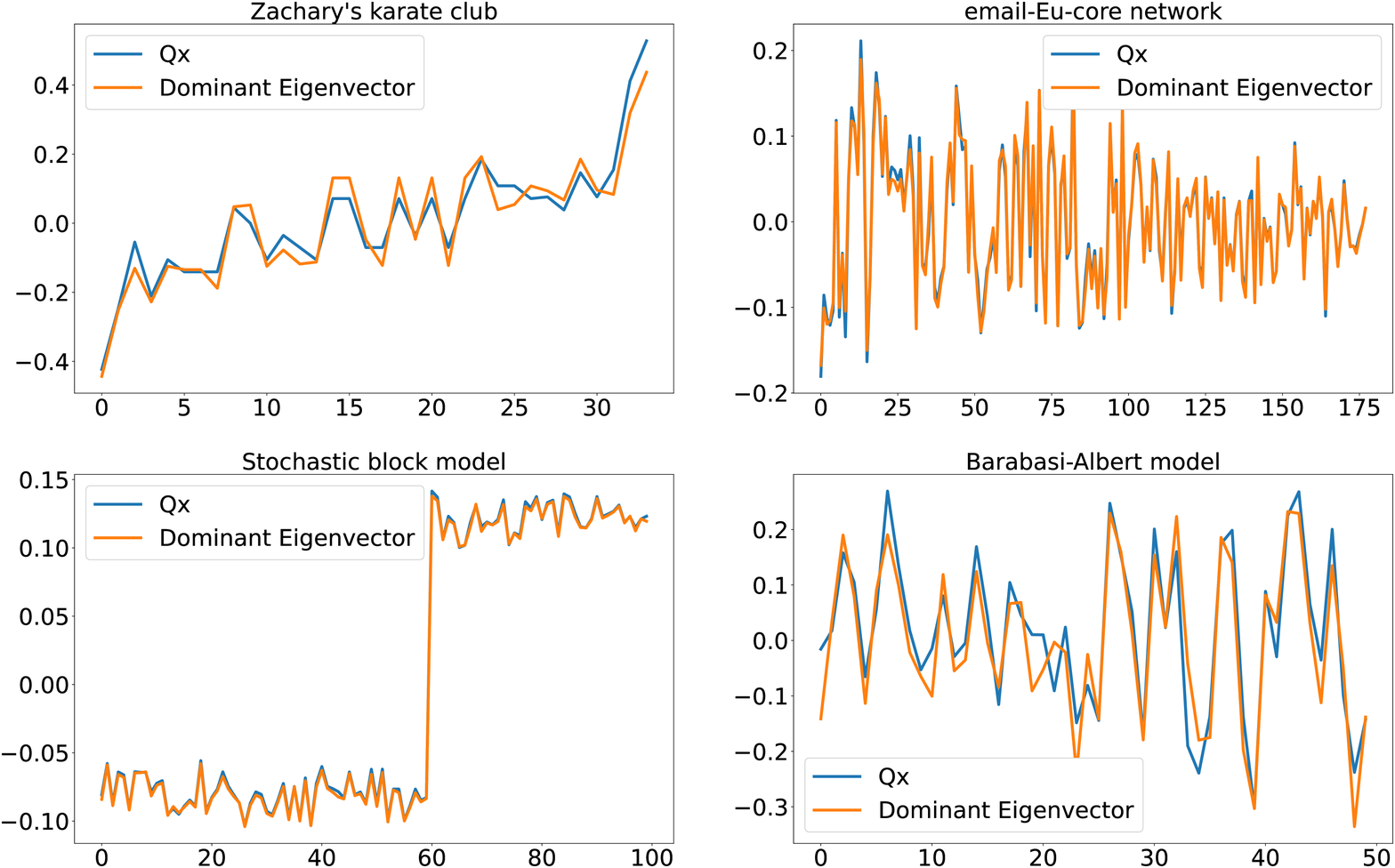}
	\caption{Comparison of the dominant eigenvector and the unit vector of $\Q x$ for four different datasets.}
	\label{fig:fourdata}
\end{figure}

In \rfig{fourdata}, we show the dominant eigenvector and the unit vector of $\Q x$ for four different datasets. These four datasets include two synthetic datasets (a stochastic block model with two blocks \cite{holland1983stochastic} and a Barab\'{a}si–Albert (BA) model \cite{barabasi1999emergence}), and two real-world datasets (the Zachary's karate club dataset \cite{zachary1977information} and the email-Eu-core dataset \cite{leskovec2007graph}). Note that we only use the subgraph in the top two communities of the email-Eu-core dataset in this figure. As shown in \rfig{fourdata}, the differences are very small and the CAFE-GCN indeed computes good approximations of the dominant eigenvectors.

\subsection{Using the CAFE-GCN for dimensionality reduction}\label{reduction}

In this section, we show how one can use the CAFE-GCN for the dimensionality reduction problem. Suppose that the $n$ data points $\{u_1, u_2, \ldots, u_n\}$ are in $\mathbb{R}^m$. Let $u_i=(u_{i,1}, u_{i,2} \ldots, u_{i,m})$. Assume that the set of $n$ points has zero-mean, i.e., for all $j=1,2, \ldots, m$, $$\sum_{i=1}^n u_{i,j}=0.$$ Note that if the set of $n$ data points do not have zero-mean, one can subtract its centroid to make it zero-mean, i.e., the $m$-vector $u_0=(u_{0,1}, u_{0,2}, \ldots, u_{0,m})$ with $$u_{0,j}=\frac{1}{n}\sum_{i=1}^n u_{i,j}.$$

Now we represent these $n$ data points by an $n \times m$ matrix $X$ with the $i^{th}$ row of $X$ being $u_i$. Consider the $n \times n$ covariance matrix $X X^T$. Then $\Q=X X^T$ is an $n \times n$ symmetric matrix with zero row sums and column sums. Moreover, it was shown in \cite{chang2019generalized} that $\Q=X X^T$ is an modularity matrix with the bivariate distribution $$p_{U,W}(u,w)=\frac{\exp(\theta ||u-w||^2)}{\sum_{w_1} \sum_{w_2} \exp(\theta ||w_1-w_2||^2)},$$ when $\theta$ is very small. One widely used method for dimensionality reduction is the principal component analysis (PCA) that finds the $K$ eigenvectors corresponding to the $K$ largest eigenvalues of $\Q$. Here we show that one can also use the CAFE-GCN in Algorithm \ref{alg:CAFE} to obtain an $n \times K$ embedding matrix $\Q H$ for the approximations of the eigenvectors from PCA. Note that the $i^{th}$ row of $\Q H$ is the embedding $K$-vector for $u_i$.

Let $W^{(\ell)}=X^T H^{(\ell)}$. Then we can rewrite \req{softmax1111b} as follows:
\beq{softmax1111d}
H^{(\ell+1)}=\sigma(X X^T H^{(\ell)})=\sigma(X W^{(\ell)}).
\eeq
Multiplying both side of \req{softmax1111d} by $X^T$ yields
\bieeeeq{reduce1111}
W^{(\ell+1)}&=&X^T H^{(\ell+1)}=X^T \sigma(X W^{(\ell)})\nonumber\\
&=&F(X, W^{(\ell)}),
\eieeeeq
where $F(X, W)=X^T \sigma(X W)$ is a function with two input matrices: an $n \times m$ matrix $X$ and $m \times K$ matrix $W$. This leads to an iterative way to find the weight matrices $W^{(\ell)}$.

Now we give the physical meaning for the matrix $W$ when the CAFE-GCN converges. Let $W_j$ be the $j^{th}$ column of $W$, i.e., $$W=[W_1 \vert W_2 \vert \ldots \vert W_K].$$ Suppose that when the CAFE-GCN converges, the matrix $H$ is a partition matrix (see Step (9) of Algorithm \ref{alg:CAFE}), i.e., every row of $H$ contains exactly one $1$ and the rest elements are $0$. Let $H_j$ be the $j^{th}$ column of $H$, and $S_j$ be the set of rows in $H_j$ that have value $1$, i.e., the set of nodes that are assigned to the $j^{th}$ cluster. Since $W=X^T H$, we have $$W_j= [u_1 \vert u_2 \vert \ldots \vert u_n] H_j =\sum_{\ell \in S_j}u_\ell.$$ This shows that $W_j$ is in line with the centroid of the set of nodes in $S_j$. As the $i^{th}$ row of $\Q H=X W$ is the embedding $K$-vector for $u_i$, the $j^{th}$ element of that embedding vector is simply the inner product of $u_i$ and $W_j$ (before the QR decomposition).

Such an interpretation is in line with the explanation of why Convolutional Neural Networks (CNN) work in the two seminal papers \cite{kuo2016understanding,kuo2017cnn}. One key insight in \cite{kuo2016understanding,kuo2017cnn} is that a CNN can be decomposed into two stages of subnetworks: the feature extraction (FE) subnet and the decision-making (DM) subnet. The FE subnet in the first stage conducts clustering and produces a new representation of a data point through a sequence of RECOS transforms in \cite{kuo2017cnn}. The DM subnet in the second stage then classifies the new representations according to decision labels. As such, a CNN basically performs two tasks: clustering in the first stage and classification in the second stage. As pointed out in \cite{kuo2017cnn}, the classification part in the second stage is similar to the multilayer perceptrons (MLPs) introduced by Rosenblatt in \cite{rosenblatt1958perceptron}, and this in general only requires a small number of layers. On the other hand, the clustering part in the first stage requires stacking more layers of RECOS transforms and it is less understood how it works. Similarly, one can also decompose a GCN into two stages of subnetworks: the feature extraction subnet and the decision-making subnet. Our CAFE-GCN basically explains the clustering part in the first stage. One key difference between \cite{kuo2017cnn} and CAFE-GCN is that the $K$-means clustering is used in \cite{kuo2017cnn} while the softmax clustering is used in CAFE-GCN. The $K$-means clustering requires knowing the number of clusters $K$ while the softmax clustering, as a modularity maximization algorithm, does not require the number of clusters in advance.

To show the effectiveness of Algorithm \ref{alg:CAFE}, we compare the embedding vectors from Algorithm \ref{alg:CAFE} with the eigenvectors from PCA. Let $\hat H_j$ be the $j^{th}$ column of the output matrix $\hat H$ from Algorithm \ref{alg:CAFE} and $v_j$ be the eigenvector corresponding to $\lambda_j$ of $\Q=X X^T$. We compute the projection of $\hat H_j$ onto the subspace spanned by $\{v_1, v_2, \ldots, v_K\}$, denoted by $P_j$, as follows:
\beq{proj1111}
P_j=\sum_{\ell=1}^K c_{j,\ell} v_\ell,
\eeq
where $c_{j,\ell} ={\hat H}_j^T v_\ell$ is the inner product of ${\hat H}_j$ and $v_{\ell}$. Note that
\beq{proj3333}
||\hat H_j- P_j||^2 \le 1- \sum_{\ell=1}^K c_{j,\ell}^2.
\eeq

If $\sum_{\ell=1}^K c_{j,\ell}^2$ is very close to $1$, then we know that $||\hat H_j- P_j||^2$ is very close to $0$ and thus $\hat H_j$ is very close to $P_j$.

\section{The multi-layer CAFE-GCN}\label{multi}
One can further stack multiple CAFE-GCNs together to form a multi-layer CAFE-GCN. By doing so, we can extract multi-resolution features of the dataset. Our approach is based on the fast unfolding algorithm in \cite{chang2017probabilistic} that serves a multi-resolution clustering algorithm. The fast unfolding algorithm in \cite{chang2017probabilistic} is a generalization of the Louvain algorithm \cite{blondel2008fast} that recursively merges nodes in clusters into supernodes to form a coarsened graph. Such a step is known as {\em graph coarsening} in \cite{hamilton2017representation}. LouvainNE \cite{bhowmick2020louvainne} uses the Louvain algorithm for node embedding. The differences between LouvainNE and the multi-layer CAFE-GCN are in two aspects: (i) the {\em pooling} step that defines how the edge weights of the coarsened graph are specified, and (ii) the embedding step that specifies how the embedding vectors are obtained from the clustering results. As mentioned in the previous section, the embedding step in CAFE-GCN can be interpreted as a step to generate good approximations of the dominant eigenvectors. However, there is no physical interpretation of the embedding from LouvainNE. In the pooling step, the bivariate distribution characterization in \cite{chang2017probabilistic} provides a natural way to generate the new bivariate distribution of the coarsened graph. On the other hand, there is no explanation for LouvainNE on how the edge weights should be generated for the coarsened graph (and the authors defer that as a future work). The two steps of graph coarsening and pooling as described for the multi-layer CAFE-GCN are described as follows:

{\bf Graph coarsening.} Instead of using the softmax assignment, one can use the hardmax assignment in Algorithm \ref{alg:CAFE}. The matrix $H$ now is a $n \times K$ partition matrix with each row indicating the cluster of the corresponding node. Let $S_i$, $i=1,2,\ldots, K$, be the set of nodes of cluster $i$. Aggregate the nodes in $S_i$ into a supernode $\tilde u_i$. Now we have a new dataset of $K$ nodes $\{\tilde u_1, \tilde u_2, \ldots, \tilde u_K\}$.

{\bf Pooling.} For the new dataset, we use the ``inherited'' bivariate distribution $\tilde p({\tilde u_i,\tilde u_j})$ as the pooling operator:
\beq{pool1111}
\tilde p({\tilde u_i,\tilde u_j})= \sum_{u \in S_i}\sum_{w \in S_j} p(u,w).
\eeq
This yields a new $K \times K$ modularity matrix $\tilde \Q=(\tilde q(\tilde u,\tilde w))$ as follows:
\beq{pool2222}
\tilde \Q= H^T \Q H.
\eeq
Then we use the new modularity matrix for the input of the next layer. The detailed steps for the multi-layer CAFE-GCN are outlined in Algorithm \ref{alg:CAFEmul}.
\begin{algorithm}[!htbp]%
	\KwIn{A modularity matrix $\Q=(q(u,w))$.}

	\KwOut{A $K \times K$ soft assignment matrix $H$ and an $K \times K$ embedding matrix $\hat H$ of data points.}

	\noindent {\bf (1)} Initially, set $i=0$, $\Q^{(0)}=\Q$, $K^{(0)}=n$, and modularity$=0$.

	\noindent {\bf (2)} Run the CAFE-GCN algorithm in Algorithm \ref{alg:CAFE} with the input modularity matrix $\Q^{(i)}$, the dimension of the embedding vector $K^{(i)}$, and the hardmax activation function (the inverse temperature $\theta=\infty$). Let $C^{(i)}$, $H^{(i)}$ and $\tilde H^{(i)}$ be its outputs.

	\noindent {\bf (3)} If $\mbox{tr}\Big ((H^{(i)})^T \Q^{(i)} H^{(i)}\Big)$ is larger than modularity, then set $K^{(i+1)}=C^{(i)}$, $\Q^{(i+1)}=(H^{(i)})^T \Q^{(i)} H^{(i)}$, and modularity to be $\mbox{tr}(H^{(i)})^T \Q^{(i)} H^{(i)}$. Increase $i$ by $1$ and repeat from Step (2).

	\noindent {\bf (4)} Output $C^{(j)}$, $H^{(j)}$ and $\tilde H^{(j)}$ for $j=1,2, \ldots, i$.
	
	\caption{The Mutli-layer CAFE-GCN Algorithm}
	\label{alg:CAFEmul}
\end{algorithm}

Though there are several recent advances in GCN that use hierarchical clustering algorithm for learning multi-resolution features of graphs (see, e.g., GraphSAGE-GCN \cite{hamilton2017inductive}, DIFFPOOL\cite{ying2018hierarchical}, and LouvainNE \cite{bhowmick2020louvainne}), the fast unfolding algorithm in \cite{chang2017probabilistic} has the following advantages:
\begin{description}
	\item[(i)] Scalability: Let $m$ be the number of nonzero elements in the bivariate distribution $p_{U,W}(u,w)$ of the sampled graph. If the sampled graph is sparse, i.e., $m=O(n)$, then the computational complexity of the fast unfolding algorithm is $O(n+m)$ for each round of training $n$ nodes in a layer. Though the number of layers is in general unknown, it is conjectured to be $O(\log n)$ \cite{chang2017probabilistic}. Thus, it is a nearly-linear time algorithm.
	\item[(ii)] Stability: the fast unfolding algorithm in \cite{chang2017probabilistic} is a modularity maximization algorithm that increases the modularity after each training. As such, the algorithm is stable as it is guaranteed to converge after a finite number of updates.
	\item[(iii)] Interpretability: the embedding vectors at each layer are approximations of dominant eigenvectors of the modularity matrix in that layer.
\end{description}

\section{The sphere-GCN algorithm}\label{sphere}
In this section, we propose another network embedding algorithm, called the sphere-GCN algorithm. Like the CAFE-GCN algorithm, the sphere-GCN algorithm is also a trace maximization algorithm that maximizes the modularity of the matrix $\Q$. The key difference between these two algorithms is the space of embedding vectors. Instead of mapping each embedding vector into a probability vector through the softmax function in the CAFE-GCN algorithm, we map each embedding vector by renormalizing it into a high-dimensional sphere in the sphere-GCN algorithm. The detailed steps for obtaining the embedding vectors are shown in Algorithm \ref{alg:Sphere}. For the sphere-GCN algorithm, we show in the following theorem that it also converges monotonically to a local optimum of the trace maximization problem. The proof of \rthe{SphereMain} is given in Appendix \ref{p_sphere}.
\begin{algorithm}[!htbp]%
	\KwIn{An $n \times n$ symmetric modularity matrix $\Q=(q(u,w))$, the dimension of the embedding vectors $K$, the influence parameter $\beta$, and initial vector representations for the $n$ nodes $\{h_u=(h_{u,1}, h_{u,2}, \ldots, h_{u,K}), u=1,2,\ldots,n\}$.}
	\KwOut{Final vector representations for the $n$ nodes $\{h_u=(h_{u,1}, h_{u,2}, \ldots, h_{u,K}), u=1,2,\ldots,n\}$.}

	\noindent {\bf (1)} For $u=1, 2, \ldots, n$

	\noindent {\bf (2)} Compute the $K$-vector $z_{u}=\sum_{w\neq u}q(w,u) h_{w}$.

	\noindent {\bf (3)} Update $h_{u} \leftarrow h_u +\beta(z_u-h_u)$.

	\noindent {\bf (4)} Renormalize ${h}_{u}$ by letting
	$$h_u\leftarrow\frac{h_{u}}{||h_{u}||}.$$

	\noindent {\bf (5)} Repeat from Step (1) until there is no further change.

	\noindent {\bf (6)} Compute the QR decomposition for the $n \times K$ matrix $\Q H$ to find an $n \times K$ matrix ${\hat H}=({\hat h}_{u,k})$ and an $n \times n$ upper triangular matrix $R$ so that ${\hat H}R=\Q H$.

	\noindent {\bf (7)} Output the embedding matrix $\hat H$.

	\caption{The Sphere-GCN Algorithm}
	\label{alg:Sphere}
\end{algorithm}

\bthe{SphereMain}
Let $H$ be the $n \times K$ matrix with its $u^{th}$ row being $h_u$. Suppose that the influence parameter $\beta$ in Algorithm \ref{alg:Sphere} is chosen to be in $[0,1]$. Then Algorithm \ref{alg:Sphere} converges monotonically to a local optimum of the following trace maximization problem:
\bieeeeq{sphere1111}
&\max_{\{h_u\}}& \mbox{tr}(H^T \Q H)\nonumber\\
&s.t.& ||h_u|| = 1, \quad u=1,2,\ldots, n.
\eieeeeq
\ethe

\section{Experimental Results}\label{sec:experi}
In this section, we evaluate our proposed GCN Algorithms on several real-world datasets. The experimental results demonstrate that our methods outperform several well-known based-line methods for both the node classification task and the link prediction task. Also, we show that CAFE-GCN is capable of performing dimensionality reduction.

\subsection{Datasets}
In this subsection, we first introduce the datasets used in the node classification task and the link prediction task. Three real-world datasets are used, which are Cora, Wiki, and ego-Facebook. The detailed descriptions of the datasets are listed as follows.
\begin{itemize}
	\item Cora\cite{sen2008collective}: The Cora dataset is a citation network that contains $2,708$ machine-learning papers as nodes and they are classified into seven classes. The Cora dataset consists of $5,429$ edges. The edges between two nodes are the citation links.
	\item Wiki\cite{yang2015network}: The Wiki dataset contains $2,405$ nodes and $17,981$ edges from $19$ classes. The nodes represent the Wikipedia articles, and the edges represent the web links between them.
	\item ego-Facebook\cite{leskovec2012learning}: This Facebook dataset was collected from survey participants using the Facebook application developed by \cite{leskovec2012learning}. This dataset consists of $10$ ego-networks and $193$ circles with $4,039$ nodes and $88,234$ edges. Each node might be in multiple circles. Such circles include common universities, sports teams, relatives, etc. For our experiments, we delete multiple edges, self-loops, and nodes that are not in the largest component of the network, and assign each node only to the largest circle. By doing so, we have $2,851$ nodes and $62,318$ edges left in the network. Moreover, each node now belongs to exactly one of the $46$ circles left in the network.
\end{itemize}

\subsection{Baseline methods}
For both node classification and link prediction, we compare the performance of the proposed GCNs with the following baseline methods:
\begin{itemize}
	\item Graph Factorization \cite{ahmed2013distributed}: Graph Factorization uses an approximate factorization of a node similarity matrix (based on the adjacency matrix). As such, Graph Factorization preserves the first-order proximity between nodes.
	\item DeepWalk \cite{perozzi2014deepwalk}: DeepWalk generalizes the concept of the Skip-gram model \cite{mikolov2013distributed} to learn node representations by viewing a node as a word, and a uniformly truncated random walk as a natural language sentence.
	\item node2vec \cite{grover2016node2vec}: node2vec also uses the same concept as DeepWalk \cite{perozzi2014deepwalk}. The critical difference between node2vec and DeepWalk is that node2vec performs a bias random walk that allows a more flexible definition of a random walk. However, a review paper \cite{khosla2019comprehensive} shows that the biased random walks of node2vec do not have any significant gains for graphs with low clustering coefficient and low reciprocity.
	\item Large-scale information network embeddings (LINE) \cite{tang2015line}: LINE was proposed for embedding a large scale network to preserve the first-order and second-order proximity between nodes.
	\item High-order Proximity Preserved Embedding (HOPE) \cite{ou2016asymmetric}: HOPE uses the general similarity measures (such as the Katz measure \cite{katz1953new}, the rooted PageRank \cite{song2009scalable} etc.) to quantify asymmetric high-orders proximity between nodes and learns the node embeddings by solving the matrix factorization problem with the generalized Singular Value Decomposition (SVD) \cite{van1976generalizing}.
	\item GraRep \cite{cao2015grarep}: GraRep computes the $k^{th}$ power of the adjacency matrix in order to capture the $k$-order proximity between nodes and uses matrix factorization techniques to construct the global representations for nodes.
	\item Structural deep network embedding (SDNE) \cite{wang2016structural}: SDNE uses the deep autoencoder with multiple non-linear layers to preserve the neighbor structures of nodes and penalize the nodes that are similar but mapped far from each other in the embedding space by Laplacian Eigenmaps \cite{belkin2002laplacian}.
\end{itemize}

\subsection{Network node classification}\label{node_class}
For the node classification task, every node in the training set is assigned with one label. The task is to predict the labels of the nodes in the testing set by using a classifier trained from the node embedding vectors in the training set and their corresponding labels. The embedding vectors of the nodes in the testing set are then put into the classifier to produce the predicted labels.

In Table \ref{clf}, we show the comparison results of the node classification task. The classifier we use in this task is a well-known classifier, called XGBoost \cite{chen2016xgboost}. We use accuracy, $F_1$ score, and area under the receiver operating characteristic curve (ROC AUC) as benchmarks. All the metrics are computed over $100$ experiments. Note that, as there exist three clusters that only contain one node in the ego-facebook dataset, the ROC AUC score is not defined in that case. We evaluate all the embedding methods using various fractions of the datasets as the training sets of the classifier ($10\%$, $30\%$, and $50\%$ as shown in the second row of Table \ref{clf}). As the embedding dimensions of the CAFE-GCN algorithm are determined by itself, all the other baseline methods use the same embedding dimensions as CAFE-GCN (so that the comparisons could be fair). Moreover, the embedding dimensions of the multi-layer CAFE-GCN and the semi-supervised CAFE-GCN are determined by the algorithms themselves. For each dataset, only semi-supervised CAFE-GCN use labels for embedding. Labels used in semi-supervised CAFE-GCN (marked with CAFE-GCN (semi-supervised) in Table \ref{clf}) are the same as those used in the classifiers. Moreover, all labels are used in CAFE-GCN (full label) in Table \ref{clf}. Also, the best scores of accuracy, $F_1$ scores, and ROC AUC scores are marked in bold.
\begin{table*}[!htbp]%
	\centering
	\resizebox{\textwidth}{!}{%
		\begin{tabular}{c|c?c|c|c?c|c|c?c|c|c}
			\Xhline{2\arrayrulewidth}
			\multicolumn{11}{c}{Node classification} \\ \hline
			\multirow{2}{*}{Datasets} & \multirow{2}{*}{Methods / Metrics} & \multicolumn{3}{c?}{10\% Training data} & \multicolumn{3}{c?}{30\% Training data} & \multicolumn{3}{c}{50\% Training data} \\ \cline{3-11}
			& & Accuracy & $F_1$ & AUC & Accuracy & $F_1$ & AUC & Accuracy & $F_1$ & AUC \\ \Xhline{2\arrayrulewidth}
			\multirow{12}{*}{Cora} & Graph Factorization & 0.353$\pm$0.013 & 0.281$\pm$0.018 & 0.580$\pm$0.009 & 0.461$\pm$0.013 & 0.400$\pm$0.017 & 0.641$\pm$0.009 & 0.514$\pm$0.016 & 0.460$\pm$0.018 & 0.673$\pm$0.010 \\ \cline{2-11}
			& DeepWalk & 0.671$\pm$0.016 & 0.642$\pm$0.019 & 0.786$\pm$0.011 & 0.773$\pm$0.012 & 0.755$\pm$0.013 & 0.850$\pm$0.008 & 0.808$\pm$0.009 & 0.793$\pm$0.010 & 0.874$\pm$0.006 \\ \cline{2-11}
			& node2vec & 0.695$\pm$0.016 & 0.665$\pm$0.020 & 0.800$\pm$0.013 & 0.775$\pm$0.010 & 0.757$\pm$0.012 & 0.853$\pm$0.008 & 0.804$\pm$0.011 & 0.789$\pm$0.012 & 0.872$\pm$0.007 \\ \cline{2-11}
			& LINE & 0.304$\pm$0.013 & 0.229$\pm$0.013 & 0.551$\pm$0.007 & 0.365$\pm$0.009 & 0.288$\pm$0.013 & 0.581$\pm$0.007 & 0.395$\pm$0.013 & 0.319$\pm$0.016 & 0.596$\pm$0.008 \\ \cline{2-11}
			& HOPE & 0.652$\pm$0.014 & 0.626$\pm$0.017 & 0.777$\pm$0.010 & 0.717$\pm$0.010 & 0.699$\pm$0.012 & 0.819$\pm$0.007 & 0.741$\pm$0.010 & 0.725$\pm$0.011 & 0.834$\pm$0.007 \\ \cline{2-11}
			& GraRep & 0.695$\pm$0.013 & 0.674$\pm$0.016 & 0.806$\pm$0.010 & 0.754$\pm$0.009 & 0.740$\pm$0.010 & 0.844$\pm$0.006 & 0.773$\pm$0.010 & 0.761$\pm$0.011 & 0.856$\pm$0.007 \\ \cline{2-11}
			& SDNE & 0.310$\pm$0.013 & 0.249$\pm$0.015 & 0.562$\pm$0.008 & 0.363$\pm$0.009 & 0.303$\pm$0.010 & 0.589$\pm$0.006 & 0.389$\pm$0.011 & 0.330$\pm$0.013 & 0.603$\pm$0.007 \\ \cline{2-11}
			& CAFE-GCN & 0.639$\pm$0.015 & 0.609$\pm$0.017 & 0.773$\pm$0.010 & 0.701$\pm$0.010 & 0.677$\pm$0.011 & 0.810$\pm$0.007 & 0.723$\pm$0.011 & 0.699$\pm$0.011 & 0.823$\pm$0.007 \\ \cline{2-11}
			& Mutli-layer CAFE-GCN & 0.650$\pm$0.014 & 0.622$\pm$0.020 & 0.780$\pm$0.011 & 0.714$\pm$0.010 & 0.690$\pm$0.011 & 0.818$\pm$0.007 & 0.741$\pm$0.011 & 0.718$\pm$0.011 & 0.834$\pm$0.007 \\ \cline{2-11}
			& Sphere-GCN & \textbf{0.735$\pm$0.014} & \textbf{0.713$\pm$0.017} & \textbf{0.829$\pm$0.011} & \textbf{0.786$\pm$0.008} & \textbf{0.769$\pm$0.010} & \textbf{0.862$\pm$0.006} & \textbf{0.809$\pm$0.010} & \textbf{0.795$\pm$0.011} & \textbf{0.878$\pm$0.007} \\ \hhline{~|=#===#===#===}
			& CAFE-GCN (semi-supervised) & 0.867$\pm$0.008 & 0.856$\pm$0.010 & 0.918$\pm$0.007 & 0.877$\pm$0.006 & 0.868$\pm$0.007 & 0.924$\pm$0.004 & 0.881$\pm$0.007 & 0.872$\pm$0.007 & 0.926$\pm$0.004 \\ \cline{2-11}
			& CAFE-GCN (full label) & 0.872$\pm$0.007 & 0.862$\pm$0.008 & 0.921$\pm$0.005 & 0.879$\pm$0.005 & 0.871$\pm$0.006 & 0.925$\pm$0.004 & 0.883$\pm$0.007 & 0.874$\pm$0.007 & 0.927$\pm$0.004 \\ \Xhline{2\arrayrulewidth}
			\multirow{12}{*}{Wiki} & Graph Factorization & 0.248$\pm$0.014 & 0.161$\pm$0.013 & 0.556$\pm$0.007 & 0.350$\pm$0.009 & 0.249$\pm$0.011 & 0.599$\pm$0.006 & 0.396$\pm$0.012 & 0.287$\pm$0.011 & 0.618$\pm$0.006 \\ \cline{2-11}
			& DeepWalk & 0.456$\pm$0.016 & 0.328$\pm$0.017 & 0.645$\pm$0.009 & 0.569$\pm$0.010 & 0.438$\pm$0.017 & 0.697$\pm$0.007 & 0.609$\pm$0.011 & 0.480$\pm$0.020 & 0.717$\pm$0.009 \\ \cline{2-11}
			& node2vec & 0.466$\pm$0.015 & 0.334$\pm$0.016 & 0.649$\pm$0.009 & 0.563$\pm$0.010 & 0.430$\pm$0.015 & 0.695$\pm$0.007 & 0.596$\pm$0.013 & 0.467$\pm$0.020 & 0.712$\pm$0.010 \\ \cline{2-11}
			& LINE & 0.307$\pm$0.014 & 0.224$\pm$0.012 & 0.587$\pm$0.006 & 0.388$\pm$0.010 & 0.285$\pm$0.009 & 0.617$\pm$0.005 & 0.419$\pm$0.012 & 0.309$\pm$0.011 & 0.628$\pm$0.006 \\ \cline{2-11}
			& HOPE & 0.447$\pm$0.016 & 0.327$\pm$0.014 & 0.644$\pm$0.007 & 0.530$\pm$0.011 & 0.392$\pm$0.010 & 0.676$\pm$0.005 & 0.553$\pm$0.010 & 0.413$\pm$0.011 & 0.687$\pm$0.005 \\ \cline{2-11}
			& GraRep & 0.512$\pm$0.015 & 0.373$\pm$0.014 & 0.670$\pm$0.008 & 0.588$\pm$0.009 & 0.443$\pm$0.013 & 0.703$\pm$0.006 & 0.612$\pm$0.012 & 0.469$\pm$0.016 & 0.716$\pm$0.007 \\ \cline{2-11}
			& SDNE & 0.281$\pm$0.013 & 0.205$\pm$0.014 & 0.578$\pm$0.006 & 0.356$\pm$0.010 & 0.267$\pm$0.010 & 0.607$\pm$0.005 & 0.387$\pm$0.012 & 0.295$\pm$0.012 & 0.620$\pm$0.005 \\ \cline{2-11}
			& CAFE-GCN & 0.447$\pm$0.015 & 0.323$\pm$0.013 & 0.645$\pm$0.008 & 0.515$\pm$0.010 & 0.381$\pm$0.013 & 0.673$\pm$0.006 & 0.539$\pm$0.012 & 0.403$\pm$0.014 & 0.683$\pm$0.007 \\ \cline{2-11}
			& Mutli-layer CAFE-GCN & 0.450$\pm$0.015 & 0.324$\pm$0.014 & 0.643$\pm$0.007 & 0.522$\pm$0.011 & 0.383$\pm$0.012 & 0.671$\pm$0.006 & 0.546$\pm$0.011 & 0.404$\pm$0.013 & 0.682$\pm$0.006 \\ \cline{2-11}
			& Sphere-GCN & \textbf{0.541$\pm$0.014} & \textbf{0.394$\pm$0.018} & \textbf{0.681$\pm$0.009} & \textbf{0.617$\pm$0.011} & \textbf{0.480$\pm$0.017} & \textbf{0.721$\pm$0.008} & \textbf{0.644$\pm$0.011} & \textbf{0.518$\pm$0.019} & \textbf{0.737$\pm$0.008} \\ \hhline{~|=#===#===#===}
			& CAFE-GCN (semi-supervised) & 0.555$\pm$0.012 & 0.409$\pm$0.016 & 0.691$\pm$0.008 & 0.658$\pm$0.009 & 0.559$\pm$0.019 & 0.765$\pm$0.012 & 0.706$\pm$0.010 & 0.644$\pm$0.020 & 0.810$\pm$0.012 \\ \cline{2-11}
			& CAFE-GCN (full label) & 0.728$\pm$0.012 & 0.600$\pm$0.026 & 0.789$\pm$0.013 & 0.766$\pm$0.008 & 0.697$\pm$0.026 & 0.838$\pm$0.015 & 0.777$\pm$0.011 & 0.725$\pm$0.025 & 0.852$\pm$0.014 \\ \Xhline{2\arrayrulewidth}
			\multirow{12}{*}{ego-Facebook} & Graph Factorization & 0.379$\pm$0.011 & 0.106$\pm$0.007 & N/A & 0.439$\pm$0.011 & 0.145$\pm$0.009 & N/A & 0.466$\pm$0.011 & 0.161$\pm$0.008 & N/A \\ \cline{2-11}
			& DeepWalk & \textbf{0.611$\pm$0.015} & \textbf{0.220$\pm$0.016} & N/A & \textbf{0.679$\pm$0.010} & 0.290$\pm$0.014 & N/A & 0.703$\pm$0.011 & 0.314$\pm$0.016 & N/A \\ \cline{2-11}
			& node2vec & 0.593$\pm$0.016 & 0.214$\pm$0.016 & N/A & 0.662$\pm$0.011 & 0.278$\pm$0.011 & N/A & 0.683$\pm$0.011 & 0.302$\pm$0.014 & N/A \\ \cline{2-11}
			& LINE & 0.494$\pm$0.013 & 0.155$\pm$0.009 & N/A & 0.543$\pm$0.009 & 0.194$\pm$0.009 & N/A & 0.562$\pm$0.010 & 0.212$\pm$0.011 & N/A \\ \cline{2-11}
			& HOPE & 0.526$\pm$0.013 & 0.175$\pm$0.010 & N/A & 0.588$\pm$0.011 & 0.225$\pm$0.011 & N/A & 0.613$\pm$0.013 & 0.249$\pm$0.015 & N/A \\ \cline{2-11}
			& GraRep & 0.540$\pm$0.013 & 0.183$\pm$0.011 & N/A & 0.605$\pm$0.009 & 0.241$\pm$0.011 & N/A & 0.630$\pm$0.012 & 0.268$\pm$0.015 & N/A \\ \cline{2-11}
			& SDNE & 0.510$\pm$0.014 & 0.172$\pm$0.012 & N/A & 0.580$\pm$0.010 & 0.233$\pm$0.011 & N/A & 0.606$\pm$0.009 & 0.259$\pm$0.011 & N/A \\ \cline{2-11}
			& CAFE-GCN & 0.558$\pm$0.015 & 0.202$\pm$0.013 & N/A & 0.651$\pm$0.011 & 0.279$\pm$0.015 & N/A & 0.683$\pm$0.011 & 0.308$\pm$0.017 & N/A \\ \cline{2-11}
			& Mutli-layer CAFE-GCN & 0.566$\pm$0.015 & 0.203$\pm$0.014 & N/A & 0.652$\pm$0.010 & 0.280$\pm$0.016 & N/A & 0.684$\pm$0.010 & 0.310$\pm$0.016 & N/A \\ \cline{2-11}
			& Sphere-GCN & 0.581$\pm$0.015 & 0.216$\pm$0.016 & N/A & 0.675$\pm$0.011 & \textbf{0.300$\pm$0.017} & N/A & \textbf{0.707$\pm$0.009} & \textbf{0.335$\pm$0.018} & N/A \\ \hhline{~|=#===#===#===}
			& CAFE-GCN (semi-supervised) & 0.629$\pm$0.013 & 0.238$\pm$0.016 & N/A & 0.707$\pm$0.009 & 0.326$\pm$0.018 & N/A & 0.734$\pm$0.009 & 0.360$\pm$0.019 & N/A \\ \cline{2-11}
			& CAFE-GCN (full label) & 0.681$\pm$0.015 & 0.275$\pm$0.018 & N/A & 0.728$\pm$0.009 & 0.344$\pm$0.014 & N/A & 0.743$\pm$0.010 & 0.374$\pm$0.021 & N/A \\ \Xhline{2\arrayrulewidth}
		\end{tabular}%
	}
	\caption{Node classification task on various experimental settings.}
	\label{clf}
\end{table*}

As shown in Table \ref{clf}, the proposed sphere-GCN has the best performance for (almost) all the experimental settings. Although DeepWalk is slightly better than sphere-GCN in the ego-facebook dataset when the ratio of the training set is low, the differences in the metrics between the two algorithms are rather small.

CAFE-GCN also has good performance for many experimental settings. Although both the CAFE-GCN and the sphere-GCN maximize the modularity to obtain approximations of the top $K$ eigenvectors, the performance of CAFE-GCN is not as good as sphere-GCN. The reason is that the embedding vectors of CAFE-GCN are only composed of positive numbers, and that might leads to different convergence results of \req{opti1111tr}. However, Table \ref{clf} also shows that CAFE-GCN (semi-supervised) could be benefited from the labeled data and have tremendous improvements in various metrics. Moreover, Table \ref{cora_clf} demonstrates that CAFE-GCN (semi-supervised) could even be as effective as the SplineCNN \cite{fey2018splinecnn}, which is the state-of-the-art method for node classification on the Cora dataset \cite{papers_with_code} without using additional side information like node attributes. The accuracy of the SplineCNN is directly based on the original paper \cite{fey2018splinecnn}, and all the experimental settings are the same as those in \cite{fey2018splinecnn}, i.e., $1,708$ nodes for training and $500$ nodes for testing.
\begin{table}[!htbp]%
	\centering
	\resizebox{\columnwidth}{!}{%
		\begin{tabular}{ccc}
			\hline
			\multicolumn{3}{c}{Node classification on Cora dataset} \\ \hline
			\multicolumn{1}{c|}{Metric / Method} & SplineCNN & \begin{tabular}[c]{@{}c@{}}CAFE-GCN\\ (semi-supervised)\end{tabular} \\ \hline
			\multicolumn{1}{c|}{Accuracy} & 89.48 $\pm$ 0.31 & 89.42 $\pm$ 1.03
		\end{tabular}%
	}
	\caption{Comparison between the state-of-the-art method and the CAFE-GCN (semi-supervised) for node classification on the Cora dataset.}
	\label{cora_clf}
\end{table}

\subsection{Network link prediction}\label{link_pred}
For the link prediction task, we aim to predict whether there is an edge between two nodes. We treat this task as a binary classification problem. We concatenate the two embedding vectors of a pair of two nodes as a data point. If there exists an edge between two nodes, the data point obtained by the concatenation of the two embedding vectors is labeled as $1$. Otherwise, it is labeled as $0$. We also use XGBoost \cite{chen2016xgboost} as the classifier and adopt accuracy and $F_1$ score as our benchmarks. All the metrics are computed over $100$ experiments. We evaluate all the embedding methods using various fractions of the datasets as the training sets. The results of the link prediction task are shown in Table \ref{link_clf}. The percentages in the second row of Table \ref{link_clf} are the fractions of data for the training set of the classifier. Also, the best scores of accuracy and $F_1$ scores are marked in bold.
\begin{table*}[!htbp]%
	\centering
	\resizebox{\textwidth}{!}{%
		\begin{tabular}{c|c?c|c?c|c?c|c}
			\Xhline{2\arrayrulewidth}
			\multicolumn{8}{c}{Link prediction} \\ \hline
			\multirow{2}{*}{Datasets} & \multirow{2}{*}{Methods / Metrics} & \multicolumn{2}{c?}{10\% Training data} & \multicolumn{2}{c?}{30\% Training data} & \multicolumn{2}{c}{50\% Training data} \\ \cline{3-8}
			& & Accuracy & F1 & Accuracy & F1 & Accuracy & F1 \\ \Xhline{2\arrayrulewidth}
			\multirow{10}{*}{Cora} & Graph Factorization & 0.709$\pm$0.021 & 0.721$\pm$0.015 & 0.759$\pm$0.026 & 0.779$\pm$0.021 & 0.759$\pm$0.028 & 0.783$\pm$0.023 \\ \cline{2-8}
			& DeepWalk & 0.738$\pm$0.034 & 0.773$\pm$0.022 & 0.751$\pm$0.027 & 0.788$\pm$0.017 & 0.755$\pm$0.027 & 0.792$\pm$0.018 \\ \cline{2-8}
			& node2vec & 0.806$\pm$0.030 & 0.828$\pm$0.023 & 0.812$\pm$0.030 & 0.824$\pm$0.024 & 0.820$\pm$0.032 & 0.838$\pm$0.026 \\ \cline{2-8}
			& LINE & 0.664$\pm$0.008 & 0.650$\pm$0.013 & 0.692$\pm$0.009 & 0.698$\pm$0.009 & 0.697$\pm$0.009 & 0.707$\pm$0.008 \\ \cline{2-8}
			& HOPE & 0.840$\pm$0.020 & 0.842$\pm$0.020 & 0.868$\pm$0.022 & 0.873$\pm$0.019 & 0.872$\pm$0.020 & 0.880$\pm$0.015 \\ \cline{2-8}
			& GraRep & 0.701$\pm$0.034 & 0.726$\pm$0.028 & 0.705$\pm$0.024 & 0.739$\pm$0.019 & 0.715$\pm$0.031 & 0.751$\pm$0.025 \\ \cline{2-8}
			& SDNE & 0.692$\pm$0.004 & 0.677$\pm$0.007 & 0.706$\pm$0.005 & 0.703$\pm$0.005 & 0.714$\pm$0.004 & 0.714$\pm$0.005 \\ \cline{2-8}
			& CAFE-GCN & 0.803$\pm$0.015 & 0.813$\pm$0.013 & 0.809$\pm$0.011 & 0.814$\pm$0.011 & 0.810$\pm$0.014 & 0.816$\pm$0.013 \\ \cline{2-8}
			& Mutli-layer CAFE-GCN & 0.802$\pm$0.015 & 0.813$\pm$0.013 & 0.810$\pm$0.015 & 0.813$\pm$0.015 & 0.812$\pm$0.012 & 0.819$\pm$0.012 \\ \cline{2-8}
			& Sphere-GCN & \textbf{0.884$\pm$0.011} & \textbf{0.882$\pm$0.012} & \textbf{0.905$\pm$0.010} & \textbf{0.904$\pm$0.010} & \textbf{0.912$\pm$0.009} & \textbf{0.912$\pm$0.009} \\ \Xhline{2\arrayrulewidth}
			\multirow{10}{*}{Wiki} & Graph Factorization & 0.743$\pm$0.016 & 0.682$\pm$0.016 & 0.772$\pm$0.014 & 0.714$\pm$0.015 & 0.783$\pm$0.010 & 0.723$\pm$0.013 \\ \cline{2-8}
			& DeepWalk & 0.732$\pm$0.029 & 0.719$\pm$0.022 & 0.738$\pm$0.030 & 0.728$\pm$0.022 & 0.739$\pm$0.028 & 0.729$\pm$0.021 \\ \cline{2-8}
			& node2vec & 0.773$\pm$0.027 & 0.742$\pm$0.023 & 0.792$\pm$0.024 & 0.764$\pm$0.021 & 0.794$\pm$0.023 & 0.766$\pm$0.021 \\ \cline{2-8}
			& LINE & 0.803$\pm$0.004 & 0.748$\pm$0.005 & 0.811$\pm$0.003 & 0.762$\pm$0.004 & 0.813$\pm$0.003 & 0.765$\pm$0.004 \\ \cline{2-8}
			& HOPE & 0.835$\pm$0.005 & 0.778$\pm$0.007 & 0.844$\pm$0.004 & 0.791$\pm$0.005 & 0.845$\pm$0.004 & 0.793$\pm$0.006 \\ \cline{2-8}
			& GraRep & 0.791$\pm$0.026 & 0.753$\pm$0.022 & 0.804$\pm$0.017 & 0.769$\pm$0.016 & 0.803$\pm$0.017 & 0.770$\pm$0.016 \\ \cline{2-8}
			& SDNE & 0.791$\pm$0.005 & 0.737$\pm$0.005 & 0.801$\pm$0.004 & 0.753$\pm$0.005 & 0.805$\pm$0.004 & 0.757$\pm$0.005 \\ \cline{2-8}
			& CAFE-GCN & 0.821$\pm$0.007 & 0.780$\pm$0.008 & 0.823$\pm$0.008 & 0.780$\pm$0.009 & 0.829$\pm$0.007 & 0.788$\pm$0.007 \\ \cline{2-8}
			& Mutli-layer CAFE-GCN & 0.841$\pm$0.007 & 0.802$\pm$0.009 & 0.858$\pm$0.006 & 0.826$\pm$0.006 & 0.866$\pm$0.005 & 0.834$\pm$0.006 \\ \cline{2-8}
			& Sphere-GCN & \textbf{0.869$\pm$0.006} & \textbf{0.835$\pm$0.008} & \textbf{0.876$\pm$0.007} & \textbf{0.846$\pm$0.009} & \textbf{0.877$\pm$0.008} & \textbf{0.847$\pm$0.010} \\ \Xhline{2\arrayrulewidth}
			\multirow{10}{*}{ego-Facebook} & Graph Factorization & 0.836$\pm$0.008 & 0.840$\pm$0.007 & 0.840$\pm$0.008 & 0.844$\pm$0.007 & 0.843$\pm$0.007 & 0.846$\pm$0.006 \\ \cline{2-8}
			& DeepWalk & \textbf{0.943$\pm$0.010} & \textbf{0.945$\pm$0.009} & \textbf{0.946$\pm$0.006} & \textbf{0.948$\pm$0.006} & \textbf{0.947$\pm$0.007} & \textbf{0.949$\pm$0.006} \\ \cline{2-8}
			& node2vec & 0.899$\pm$0.025 & 0.907$\pm$0.021 & 0.904$\pm$0.032 & 0.911$\pm$0.027 & 0.907$\pm$0.024 & 0.914$\pm$0.020 \\ \cline{2-8}
			& LINE & 0.909$\pm$0.004 & 0.907$\pm$0.004 & 0.910$\pm$0.005 & 0.908$\pm$0.005 & 0.911$\pm$0.005 & 0.909$\pm$0.005 \\ \cline{2-8}
			& HOPE & 0.926$\pm$0.009 & 0.927$\pm$0.010 & 0.930$\pm$0.006 & 0.932$\pm$0.006 & 0.930$\pm$0.006 & 0.932$\pm$0.006 \\ \cline{2-8}
			& GraRep & 0.909$\pm$0.014 & 0.915$\pm$0.012 & 0.911$\pm$0.013 & 0.916$\pm$0.011 & 0.914$\pm$0.011 & 0.919$\pm$0.010 \\ \cline{2-8}
			& SDNE & 0.900$\pm$0.007 & 0.901$\pm$0.006 & 0.901$\pm$0.008 & 0.903$\pm$0.007 & 0.902$\pm$0.007 & 0.903$\pm$0.006 \\ \cline{2-8}
			& CAFE-GCN & 0.876$\pm$0.009 & 0.878$\pm$0.008 & 0.878$\pm$0.010 & 0.880$\pm$0.009 & 0.890$\pm$0.010 & 0.892$\pm$0.010 \\ \cline{2-8}
			& Mutli-layer CAFE-GCN & 0.887$\pm$0.010 & 0.888$\pm$0.009 & 0.889$\pm$0.011 & 0.891$\pm$0.010 & 0.904$\pm$0.010 & 0.905$\pm$0.010 \\ \cline{2-8}
			& Sphere-GCN & 0.936$\pm$0.004 & 0.937$\pm$0.004 & 0.939$\pm$0.004 & 0.940$\pm$0.004 & 0.940$\pm$0.006 & 0.941$\pm$0.006 \\ \Xhline{2\arrayrulewidth}
		\end{tabular}%
	}
	\caption{Link prediction task on various experimental settings.}
	\label{link_clf}
\end{table*}

As in the previous task, sphere-GCN outperforms the other methods in almost all the experimental settings, except for the ego-facebook dataset. The reason is that the performance of the link prediction task for a particular dataset might depend on the similarity matrix used for generating the embedding vectors. From this experiment, it seems that DeepWalk, which chooses the ``pointwise mutual information'' as the similarity matrix \cite{hamilton2017representation}, is more suitable for the link prediction task for the ego-facebook dataset. However, the sphere-GCN that uses ``modularity'' as the similarity matrix actually achieves similar performance to DeepWalk.

\subsection{Point cloud image reconstruction}\label{image_recon}
To illustrate that our CAFE-GCN algorithm in Algorithm \ref{alg:CAFE} is able to perform dimensionality reduction well, we use the embedding vectors obtained from our algorithm to reconstruct several point cloud images that are transformed into a very high dimensional Euclidean space (by some measure-preserving transformations).

\begin{figure}[!htbp]%
		\centering
		\subfloat[Two concentric circles]{{\includegraphics[width=0.5\columnwidth]{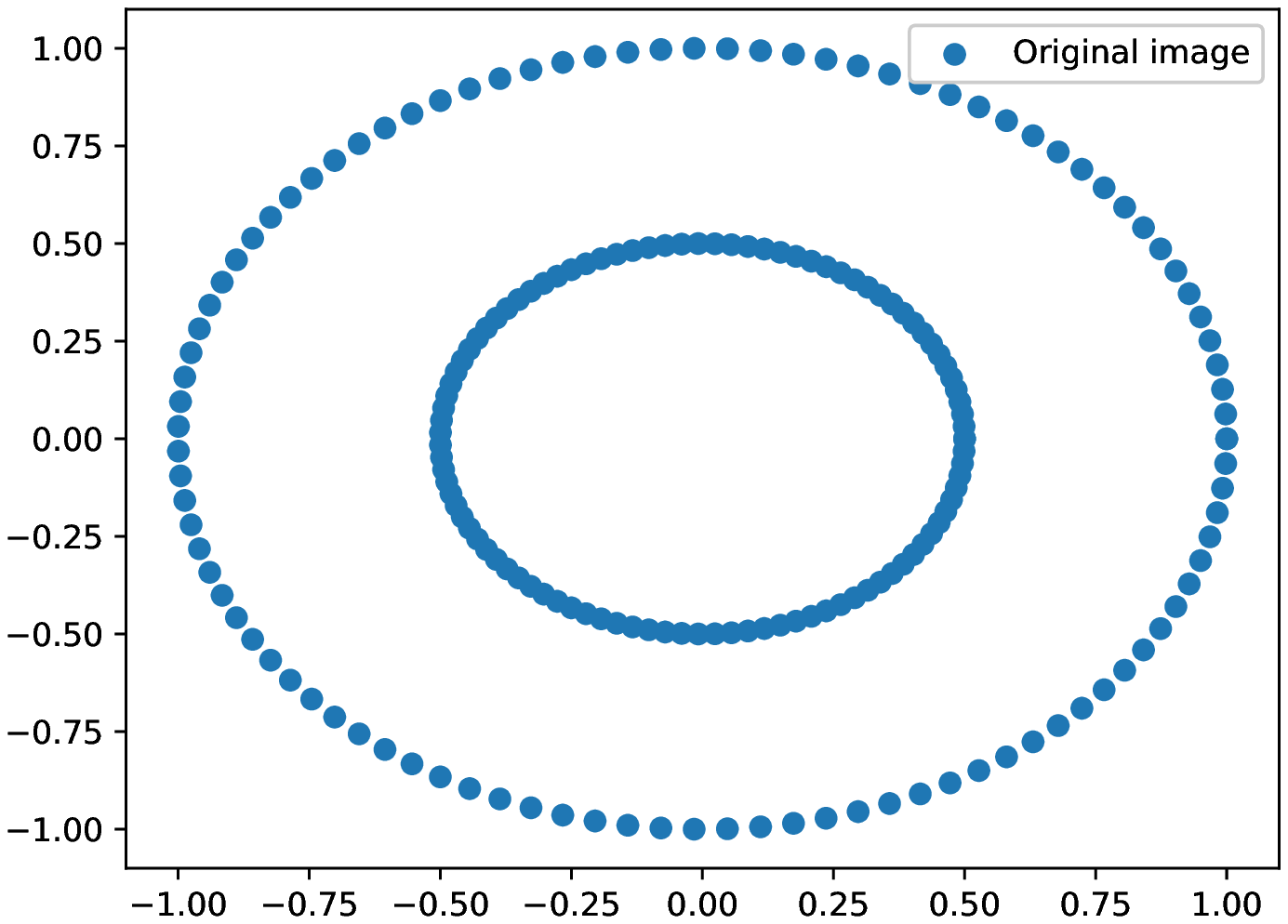}}}%
		\subfloat[Bunny]{{\includegraphics[width=0.5\columnwidth]{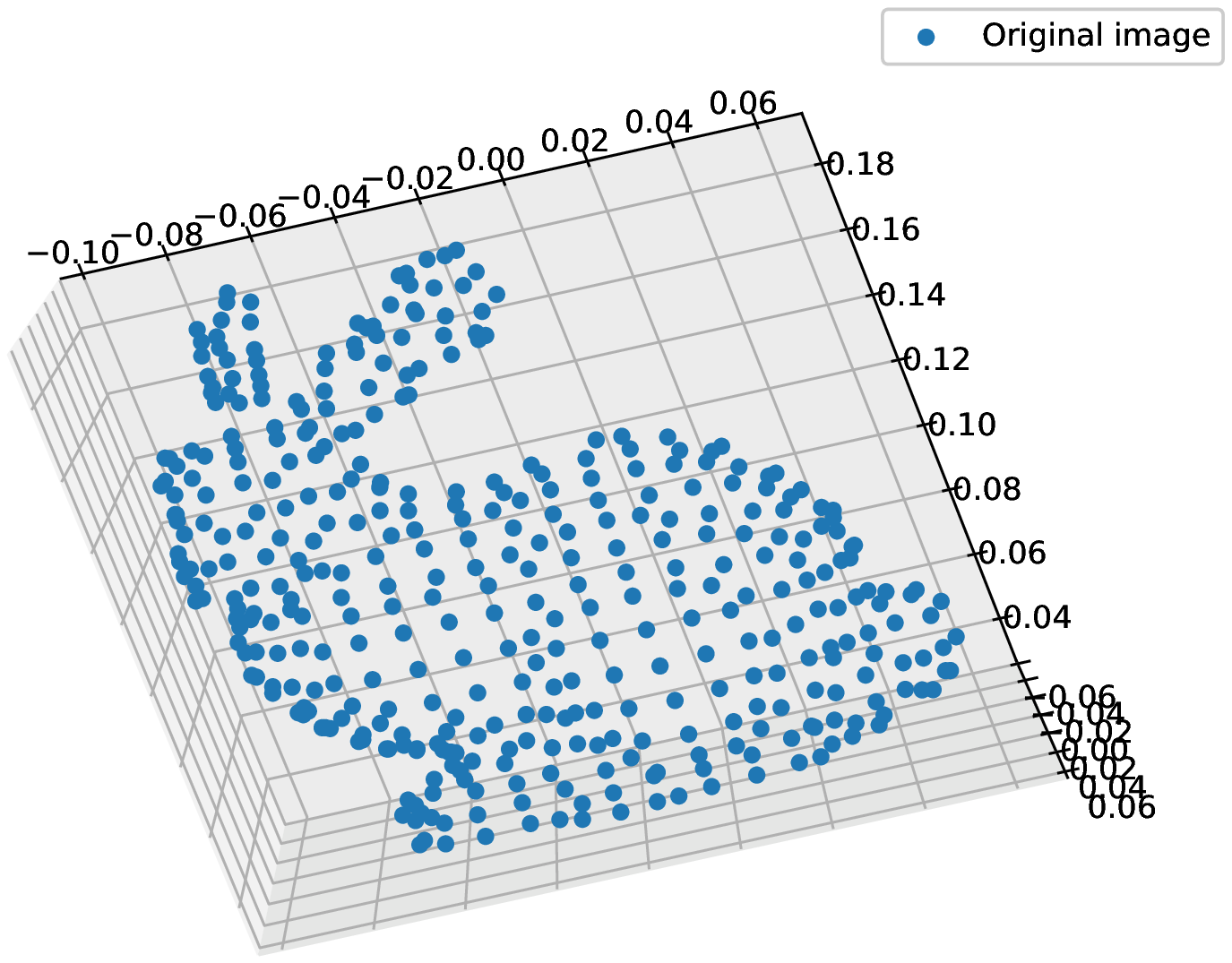}}}\\%
		\subfloat[Teapot]{{\includegraphics[width=0.5\columnwidth]{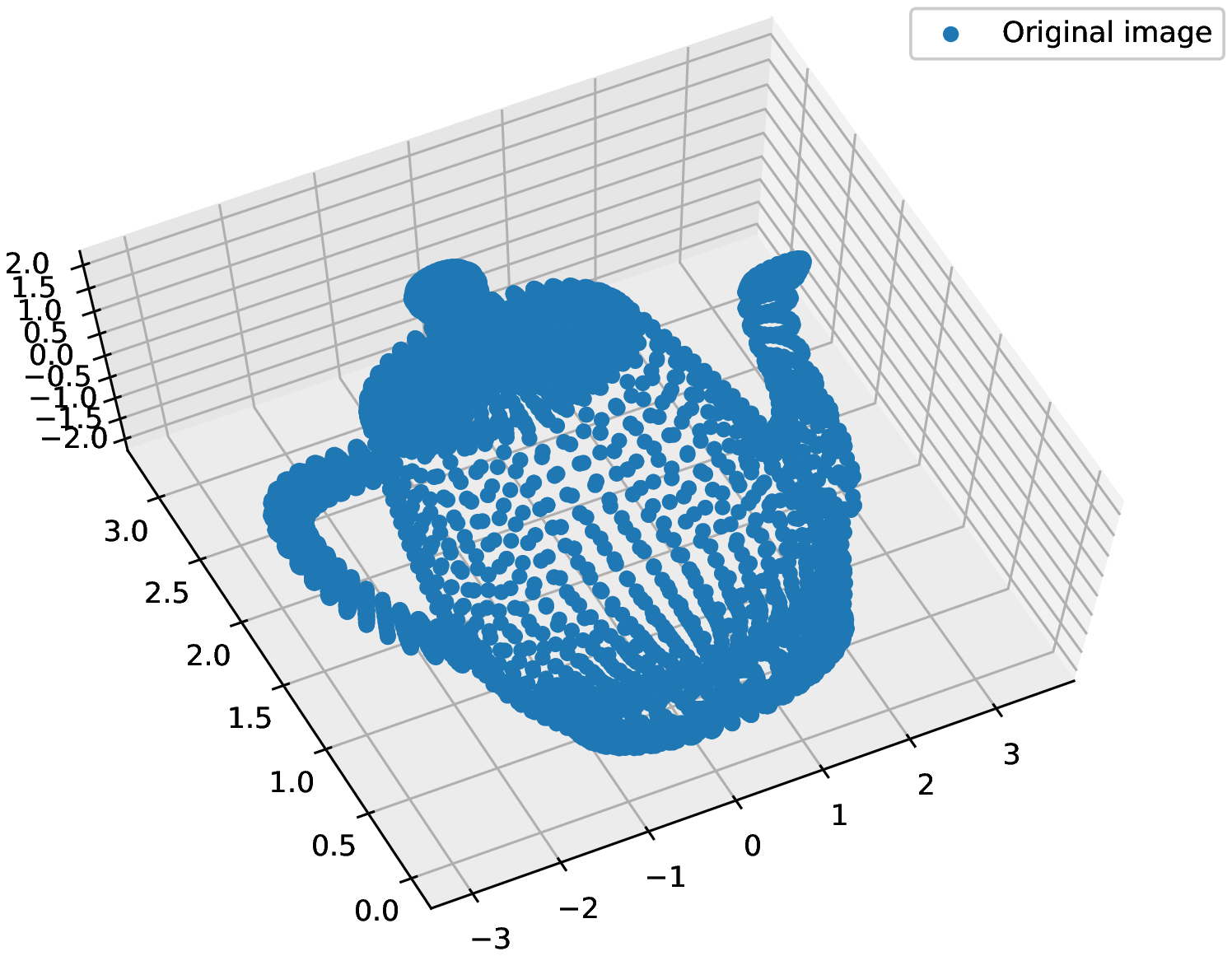}}}%
		\subfloat[Junction]{{\includegraphics[width=0.5\columnwidth]{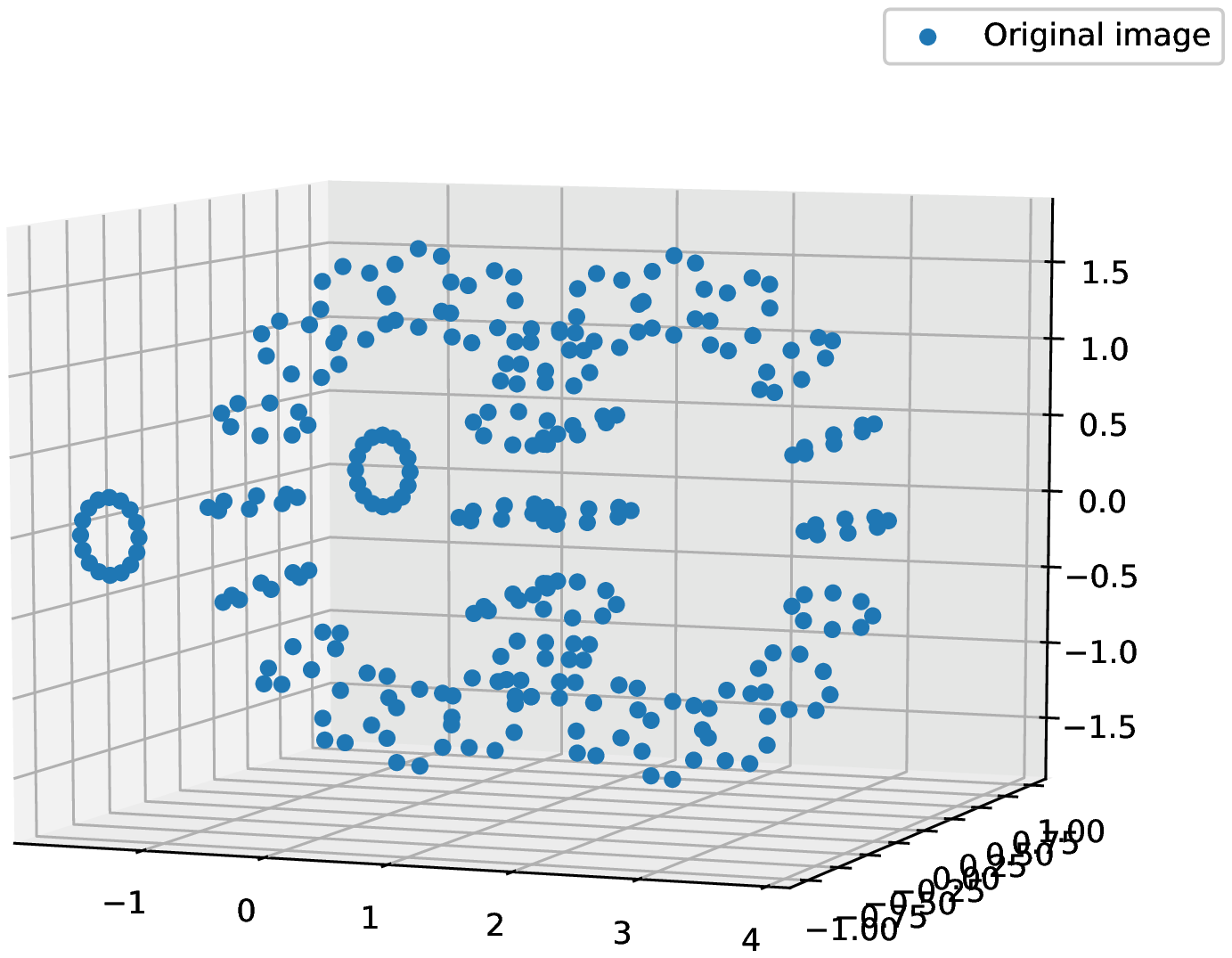}}}%
		\caption{Original point cloud images}%
		\label{fig:img_original}%
	\end{figure}

Specifically, we choose $4$ point cloud images, which are two concentric circles in 2D, Bunny in 3D \cite{Rusu_ICRA2011_PCL}, Teapot in 3D \cite{joubert_andrews_2010}, and Junction in 3D \cite{joubert_andrews_2010} as our datasets (see Figure \ref{fig:img_original}). Each point cloud image contains $n=200$, $397$, $3,644$, and $288$ data points, respectively. Then we multiply the original data points by a $3 \times L$ (resp. $2 \times L$) orthogonal matrices for the 3D (resp. 2D) point cloud images and represent these transformed data points by an $n \times L$ matrix. By doing so, each data point is transformed into an $L$-vector. In this experiment, we set $L$ to be $30$. We then subtract the column mean for each column of the $n \times L$ matrix to obtain another $n \times L$ matrix $X$ that has zero column sums. Then the $n \times n$ matrix $\Q=X X^T$ is a symmetric matrix with zero row sums and column sums. We then use our CAFE-GCN algorithm in Algorithm \ref{alg:CAFE} with the input matrix $\Q$ to compute the $K$-dimensional embedding vectors of the $n$ data points $\{{\hat h}_{u,k},u=1,2,\ldots,n,\;k=1,2,\ldots, K\}$. Instead of removing the zero columns in $H$ in Step (2) of Algorithm \ref{alg:CAFE}, we keep all the columns to evaluate how similar the space spanned by all embedding vectors from Algorithm \ref{alg:CAFE} and the space spanned by the eigenvectors are by calculating the square of the difference between the two vectors in \req{proj3333} in Table \ref{tab:cos_sim}. For the other inputs of Algorithm \ref{alg:CAFE}, we set the dimension of an embedding vector $K = 6$, and the inverse temperature $\theta = 0.010$.
\begin{table}[!htbp]%
	\centering
	\resizebox{\columnwidth}{!}{
		\begin{tabular}{|c|c|c|c|c|c|c|}
			\hline
			\multirow{2}{*}{Point Cloud Image}    & \multicolumn{6}{c|}{$1- \displaystyle\sum_{\ell=1}^K c_{j,\ell}^2$ in \req{proj3333}} \\ \cline{2-7}
			& $j = 1$  & $j = 2$  & $j = 3$  & $j = 4$ & $j = 5$ & $j = 6$ \\ \hline
			2D Two concentric circles & \bf0        & \bf0        & 0.5407   & 0.9492  & 0.9798  & 0.9800  \\ \hline
			3D Bunny                  & \bf0        & \bf0        & \bf0        & 0.8612  & 0.9208  & 0.9644  \\ \hline
			3D Teapot                 & \bf0        & \bf0        & \bf0        & 0.9392  & 0.9945  & 0.9994  \\ \hline
			3D Junction                   & \bf0        & \bf0        & \bf0        & 0.9827  & 0.4808  & 0.9043  \\ \hline
	\end{tabular}}
	\caption{Comparison of the embedding vectors from Algorithm \ref{alg:CAFE} with the eigenvectors.}
	\label{tab:cos_sim}
\end{table}

From Table \ref{tab:cos_sim}, we observe that there are three (resp. two) column vectors $\hat{H}_j$s that are very close to their projections $P_j$s into the space spanned by the three (resp. two) dominant eigenvectors of the $n \times n$ matrix $\Q$ in the 3D (resp. 2D) point cloud images. Note that the dominant eigenvectors of $\Q=X X^T$ are the results of PCA for dimensionality reduction. This means the output of our CAFE-GCN in Algorithm \ref{alg:CAFE} is able to compute the approximations of the dominant eigenvectors well and thus can also be used for dimensionality reduction. In fact, our numeral results (not shown in this paper) show that sphere-GCN is also able to perform the dimensionality reduction task.

\begin{figure}[!htbp]%
	\centering
	\subfloat[Two concentric circles]{{\includegraphics[width=0.5\columnwidth]{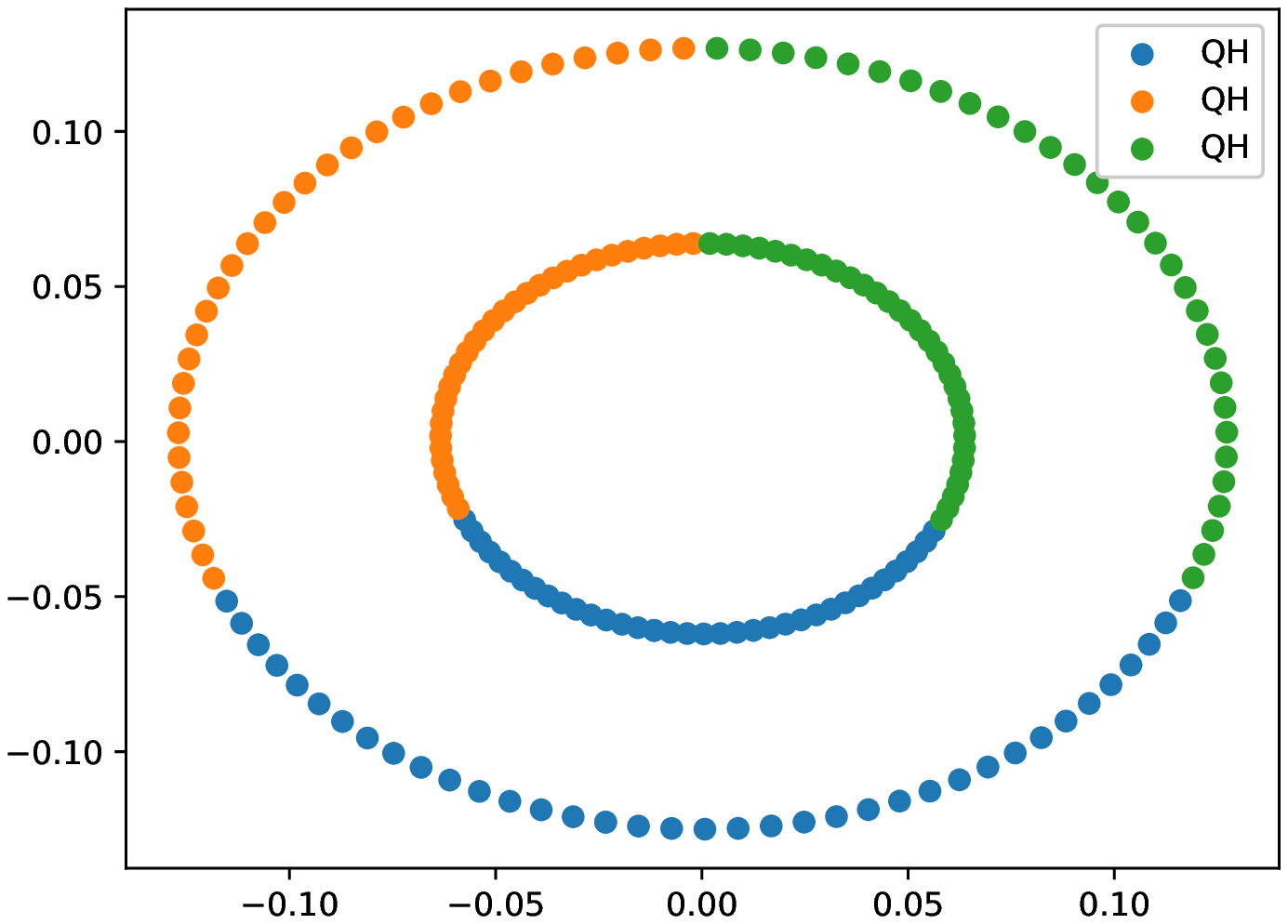}}}%
	\subfloat[Bunny]{{\includegraphics[width=0.5\columnwidth]{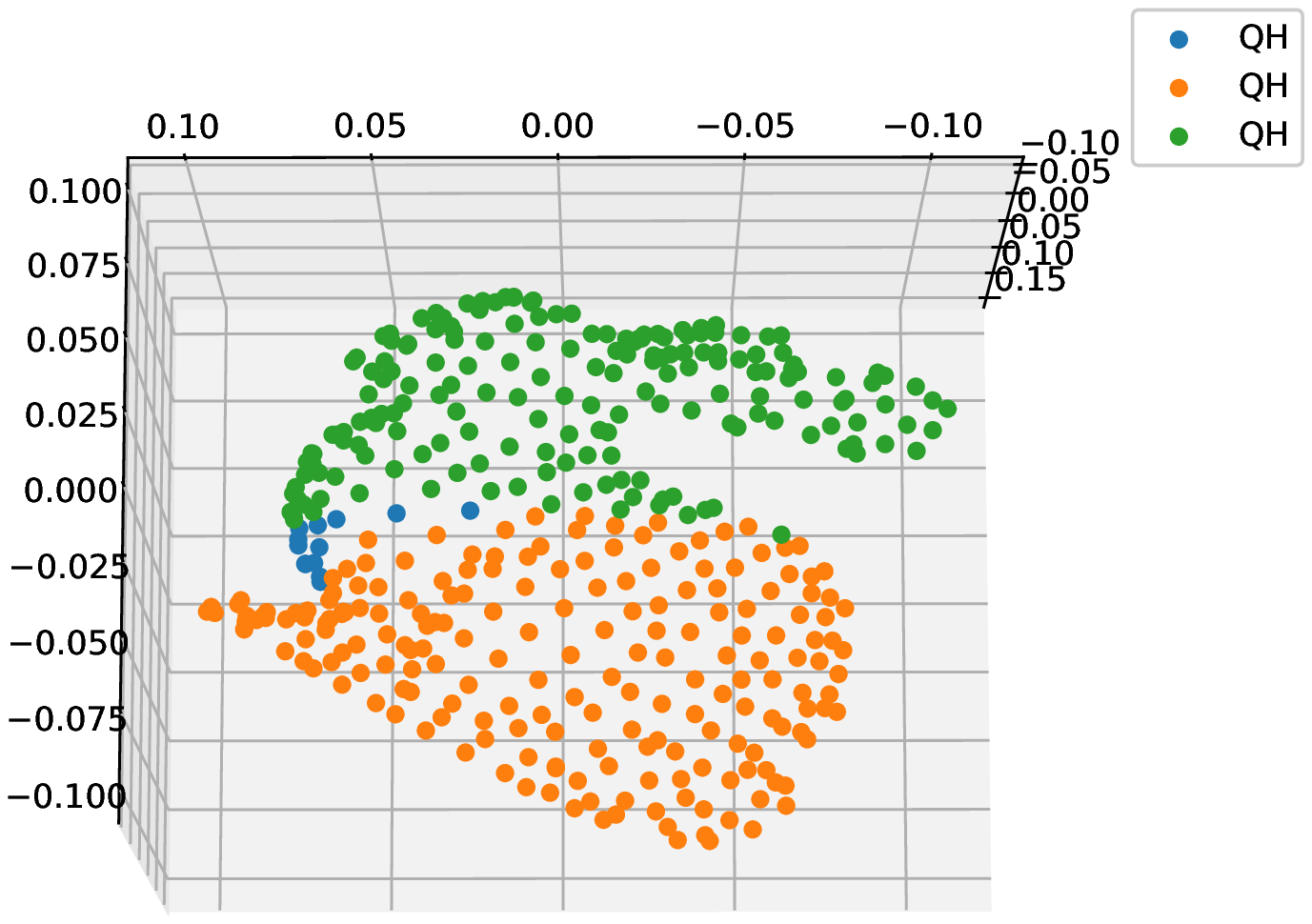}}}\\%
	\subfloat[Teapot]{{\includegraphics[width=0.5\columnwidth]{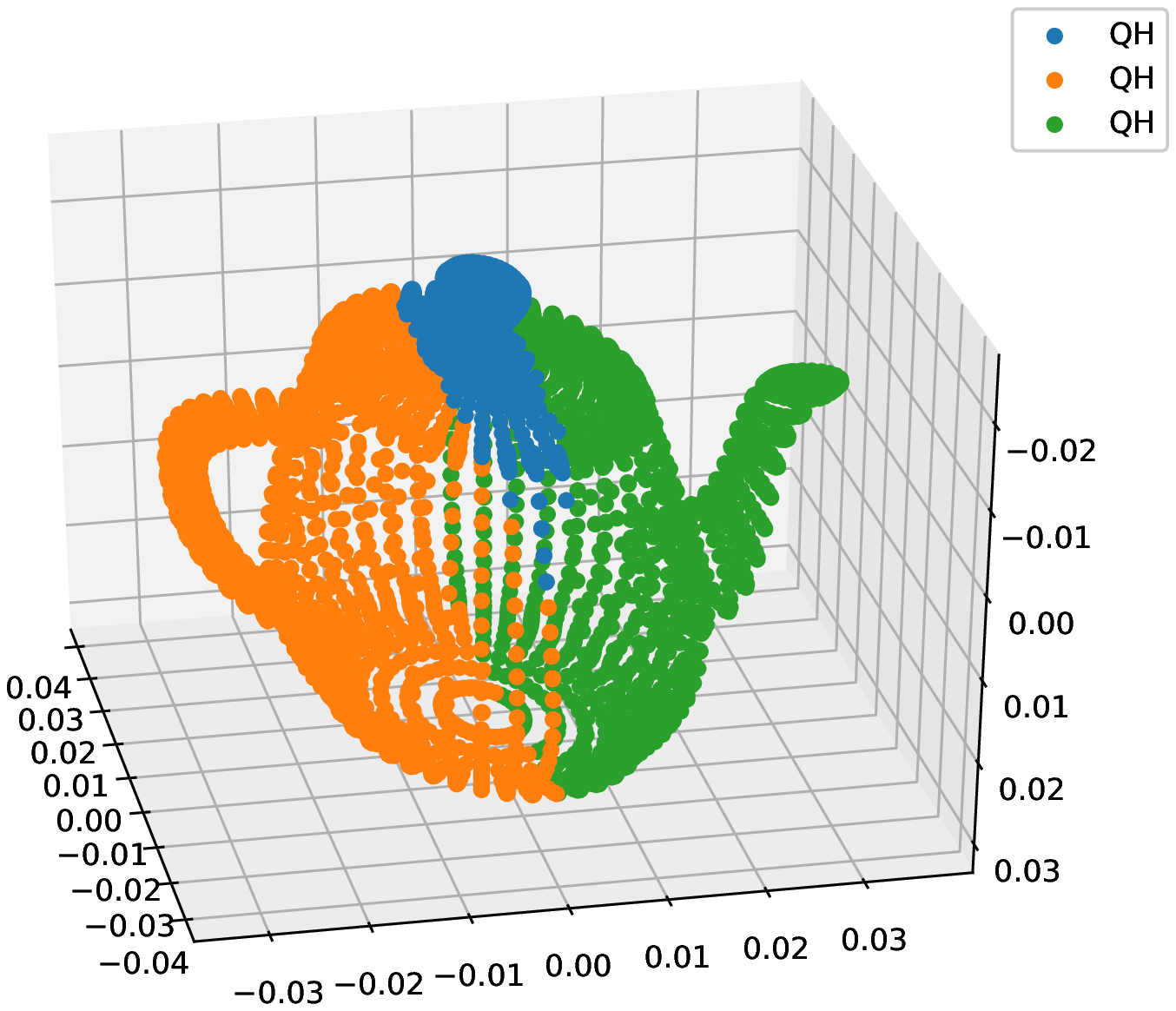}}}%
	\subfloat[Junction]{{\includegraphics[width=0.5\columnwidth]{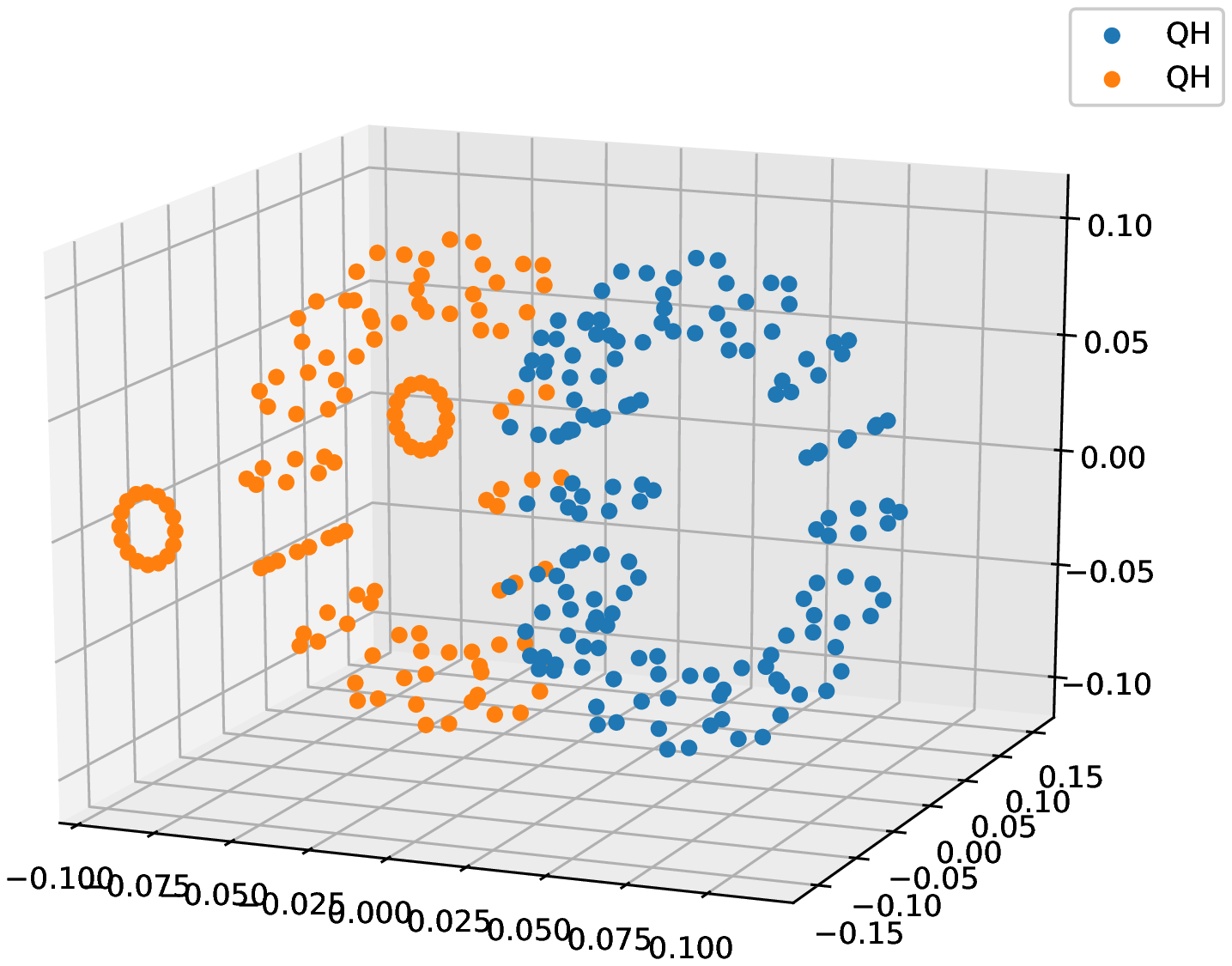}}}%
	\caption{Embedded point cloud images}%
	\label{fig:img_embed}%
\end{figure}

To show the effectiveness of Algorithm \ref{alg:CAFE}, we select the column vectors from the output matrix $\hat H$ that correspond to the closest projections into the space spanned by the dominant eigenvectors and plot the selected vectors. By doing so, the dimension of embedding vectors for 3D point cloud images (resp. 2D point cloud images) is three (resp. two). Figure \ref{fig:img_embed} shows that the reconstructed point cloud images are almost the same as the original point cloud images. In order to visualize the clustering results, we apply a hard assignment on the soft assignment matrix $H$, which is the output of the Algorithm \ref{alg:CAFE} and obtain a partition matrix consisting of only $0$ and $1$. The color of a data point represents the group it belongs to.

Looking deeper into the Junction dataset in Table \ref{tab:cos_sim} and Figure \ref{fig:img_embed}, we find that Algorithm \ref{alg:CAFE} is able to reconstruct the point cloud image and obtain good approximations of the top three dominant eigenvectors of $\Q=X X^T$. However, we only obtain two clusters from the partition matrix, which is different from the number of clusters $C=3$ in Step (4) of Algorithm \ref{alg:CAFE}. If we replace the soft assignment $H$ to the partition matrix in Step (3) of Algorithm \ref{alg:CAFE}, we can only get two dominant eigenvectors of $\Q$. The reason is that all the none zero elements of $H$ are essential, no matter how small their probabilities are. We note that the hard assignment that outputs a partition matrix consisting of only $0$ and $1$ (colors of data points in Figure \ref{fig:img_embed}) discards important information of the number of linearly independent embedding vectors, especially for nodes or data points that are difficult to cluster (as those nodes exist small probabilities of being in another cluster). To explain this, suppose that the softmax function (soft assignment) outputs three linearly independent vectors $v_1,v_2,$ and $\eta v_3$, where $\eta$ is very small after lots of iterations of softmax updates. Then using the hardmax function (hard assignment) forces $\eta$ to be $0$. This leads to a result with only two linearly independent vectors $v_1$ and $v_2$, and the dimension is reduced from $3$ to $2$. As such, it is critical to keep the third linearly independent vector even though its coefficient is very small. For a community detection algorithm, what we care about is to which community a node is most likely assigned, and it is fine to use the hard assignment that sets elements with small probabilities to be zero. However, for the network embedding problem and the dimensionality reduction problem, it is crucial to keep most of the information by using the softmax function.

\section{Conclusion}\label{conclusion}
Based on the equivalence of the network embedding problem, the trace maximization problem, and the matrix factorization problem in a sampled graph, we proposed two explainable, scalable, and stable GCN algorithms for learning graph representations: (i) CAFE-GCN and (ii) sphere-GCN. We showed that both algorithms converge monotonically to a local optimum of the trace maximization problem in a sample graph, and thus yield good approximations of the dominant eigenvectors of the modularity matrix. The key difference between these two algorithms is the space of embedding vectors. CAFE-GCN maps each embedding vector into a probability vector through the softmax function, while sphere-GCN maps each embedding vector into a unit sphere. Both proposed GCNs are {\em local} methods as they only require local information from the node itself. As such, there are linear-time implementations for our proposed GCNs when the graph is sparse. In comparison with the proposed GCN methods, the power method is a {\em global} method that requires information from all the $n$ nodes for renormalization back to a unit $n$-vector. As such, our GCN approaches are more scalable for large $n$. In addition to solving the network embedding problem, both proposed GCNs are capable of performing dimensionality reduction.

Various experiments were conducted to evaluate our proposed GCNs. Our numerical results show that sphere-GCN outperforms almost all the baseline methods for node classification and link prediction tasks. Moreover, semi-supervised CAFE-GCN could be benefited from the labeled data and have tremendous improvements in various metrics. In particular, for the node classification task on the Cora dataset \cite{papers_with_code}, CAFE-GCN (semi-supervised) achieves almost the same accuracy as the SplineCNN \cite{fey2018splinecnn}, which is the state-of-the-art method without using any side information.



%

\appendices
\section{Proof of \rthe{max}}\label{p_max}
\quad We first show \req{max1111}. Since $\{v_1, v_2, \ldots, v_n\}$ is an orthonormal basis, the $n$-vector $x$ can be represented as
\beq{max2222}
x=\sum_{i=1}^n c_i v_i,
\eeq
where $c_i=x^T v_i$ is the $i^{th}$ coordinate with respect to the orthonormal basis $\{v_1, v_2, \ldots, v_n\}$. Since $x^T x=1$, we have from \req{max2222} that
\beq{max2255}
\sum_{i=1}^n c_i^2=1.
\eeq
Using the fact that $v_i$'s are eigenvectors of $\Q$ yields
\bieeeeq{max3333}
x^T \Q x&=&\sum_{i=1}^n \lambda_i c_i^2\nonumber\\
& \le & \lambda_1 c_1^2 + \sum_{i=2}^n |\lambda_i| c_i^2.
\eieeeeq
From \req{gap1111} and \req{max2255}, it follows that
\bieeeeq{max3355}
x^T \Q x &\le & \lambda_1 c_1^2 +\deltaone \lambda_1 \sum_{i=2}^n c_i^2\nonumber\\
&=& \lambda_1((1-\deltaone) c_1^2 +\deltaone).
\eieeeeq
In conjunction with \req{max1100}, we then have
\beq{max3366}
COS(x,v_1)=c_1 \ge \sqrt{\frac{1-\epsilon-\deltaone}{1-\deltaone}}.
\eeq

Now we show \req{max5555}. Note that
\bieeeeq{max7777}
COS(v_1, \Q x)&=& v_1^T \frac{\Q x}{\sqrt{(\Q x)^T \Q x}}\nonumber\\
&=& \frac{\lambda_1 v_1^T x}{\sqrt{(\Q v)^T \Q x}}\nonumber\\
&=&\frac{\lambda_1 \cdot COS(v_1, x)}{\sqrt{(\Q x)^T \Q x}} .
\eieeeeq
Analogous to the argument in \req{max3333} and \req{max3355}, we have from the symmetry of $\Q$ that
\bieeeeq{max3355b}
(\Q x)^T \Q x&=&x^T \Q^2 x\nonumber\\
&\le& \lambda_1^2((1-\deltaone^2) c_1^2 +\deltaone^2)\nonumber\\
&=& \lambda_1^2((1-\deltaone^2) (COS(v_1, x))^2 +\deltaone^2)\label{eq:max3366a}\\
&\le& \lambda_1^2.\label{eq:max3366b}
\eieeeeq
Using \req{max3366b} in \req{max7777} yields $$COS(v_1, \Q x) \ge COS(v_1, x).$$ Moreover, using \req{max3366a} in \req{max7777} yields
\beq{max8888}
COS(v_1, \Q x) \ge \frac{COS(v_1, x)}{\sqrt{(1-\deltaone^2) (COS(v_1, x))^2 +\deltaone^2}}.
\eeq
It is straightforward to verify that the function $$f(t)=\frac{t}{\sqrt{(1-\deltaone^2) t^2 +\deltaone^2}}=\frac{1}{\sqrt{(1-\deltaone^2) +\frac{\deltaone^2}{t^2}}}$$ is increasing in $t$ for $t >0$. Using the lower bound for $COS(v_1, x)$ in \req{max1111} yields the lower bound for $COS(v_1, \Q x)$ in \req{max5555}.

\section{Proof of \rthe{SphereMain}}\label{p_sphere}
First, we note that
\bieeeeq{sphere2222}
&&\mbox{tr}(H^T \Q H)= \sum_{k=1}^K\sum_{u=1}^n\sum_{w=1}^n q(u,w)h_{u,k}h_{w,k} \nonumber\\
&&=\sum_{u=1}^n\sum_{w=1}^n q(u,w)h_{u}\cdot h_{w} \nonumber\\
&&=\sum_{u=1}^n q(u,u)|| h_u ||^2 +\sum_{u=1}^n h_{u}\cdot (\sum_{w\ne u }^n q(u,w) h_{w}) \nonumber\\
&&=\sum_{u=1}^n q(u,u)|| h_u ||^2 +\sum_{u=1}^n h_{u}\cdot z_u,
\eieeeeq
where
$$z_{u}=\sum_{w\neq u}q(w,u) h_{w}.$$ As $||h_u||^2=1$ for all $u$, $$\sum_{u=1}^n q(u,u)|| h_u ||^2=\mbox{tr}( \Q ),$$ which is a constant. Thus, it suffices to show that $$\sum_{u=1}^n h_{u}\cdot z_u$$ is monotonically increasing after each iteration. Suppose that node $u$ is updated in Step (3) of Algorithm \ref{alg:Sphere}. Let $h_u^\star=h_u +\beta(z_u-h_u)$ after the update in Step (3) and $h_u^+=h_u^\star/||h_u^\star||$ after the re-normalization in Step (4). We will show that
\beq{sphere3333}
h_{u}\cdot z_u \le h_{u}^+\cdot z_u.
\eeq
Since $$h_u^\star =(1-\beta) h_u + \beta z_u,$$ the vector $h_u^\star$ is a convex combination of the two vectors $h_u$ and $z_u$ for $0 \le \beta \le 1$, i.e., $h_u^\star$ is in the segment between $h_u$ and $z_u$. Thus, the angle between $h_u^\star$ and $z_u$ is not larger than the angle between $h_u$ and $z_u$ for $0 \le \beta \le 1$. This implies that $$h_u^+ \cdot \frac{z_u}{||z_u||} \ge h_{u}\cdot \frac{z_u}{||z_u||}$$ and thus the inequality in \req{sphere3333} holds.

\ifCLASSOPTIONcaptionsoff
  \newpage
\fi

\newpage


\bibliographystyle{IEEEtran}
\bibliography{TNNLS_CAFE_arXiv}
%

%

\begin{IEEEbiography}[{\includegraphics[width=1in,height=1.25in,clip,keepaspectratio]{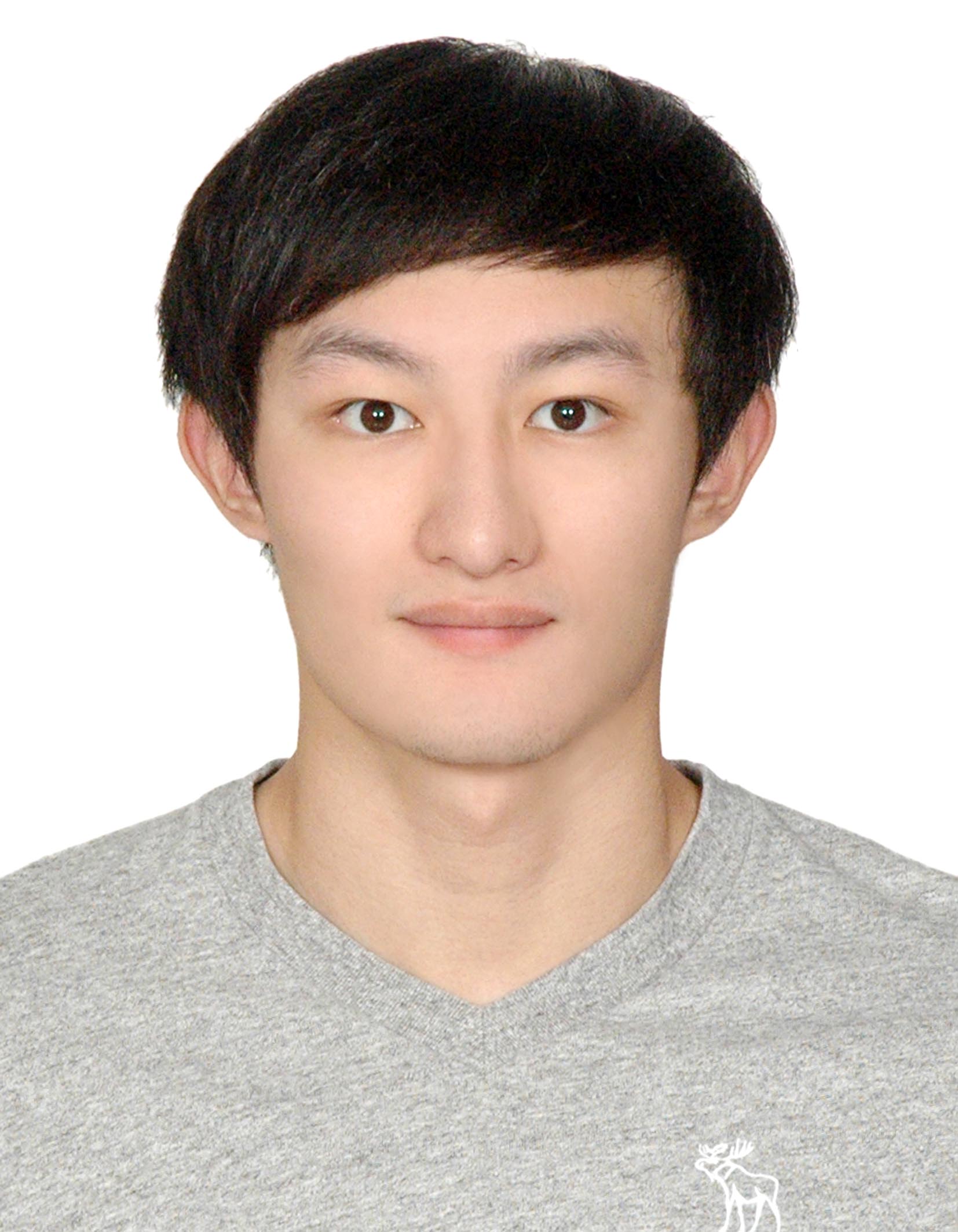}}]{Ping-En Lu} (GS'17) received his B.S. degree in communication engineering from the Yuan Ze University, Taoyuan, Taiwan (R.O.C.), in 2015. He is currently pursuing the Ph.D. degree in the Institute of Communications Engineering, National Tsing Hua University, Hsinchu, Taiwan (R.O.C.). He won the ACM Multimedia 2017 Social Media Prediction (SMP) Challenge with his team in 2017. His research interest is in network science, efficient clustering algorithms, network embedding, and deep learning algorithms. He is an IEEE Graduate Student Member.
\end{IEEEbiography}

\begin{IEEEbiography}[{\includegraphics[width=1in,height=1.25in,clip,keepaspectratio]{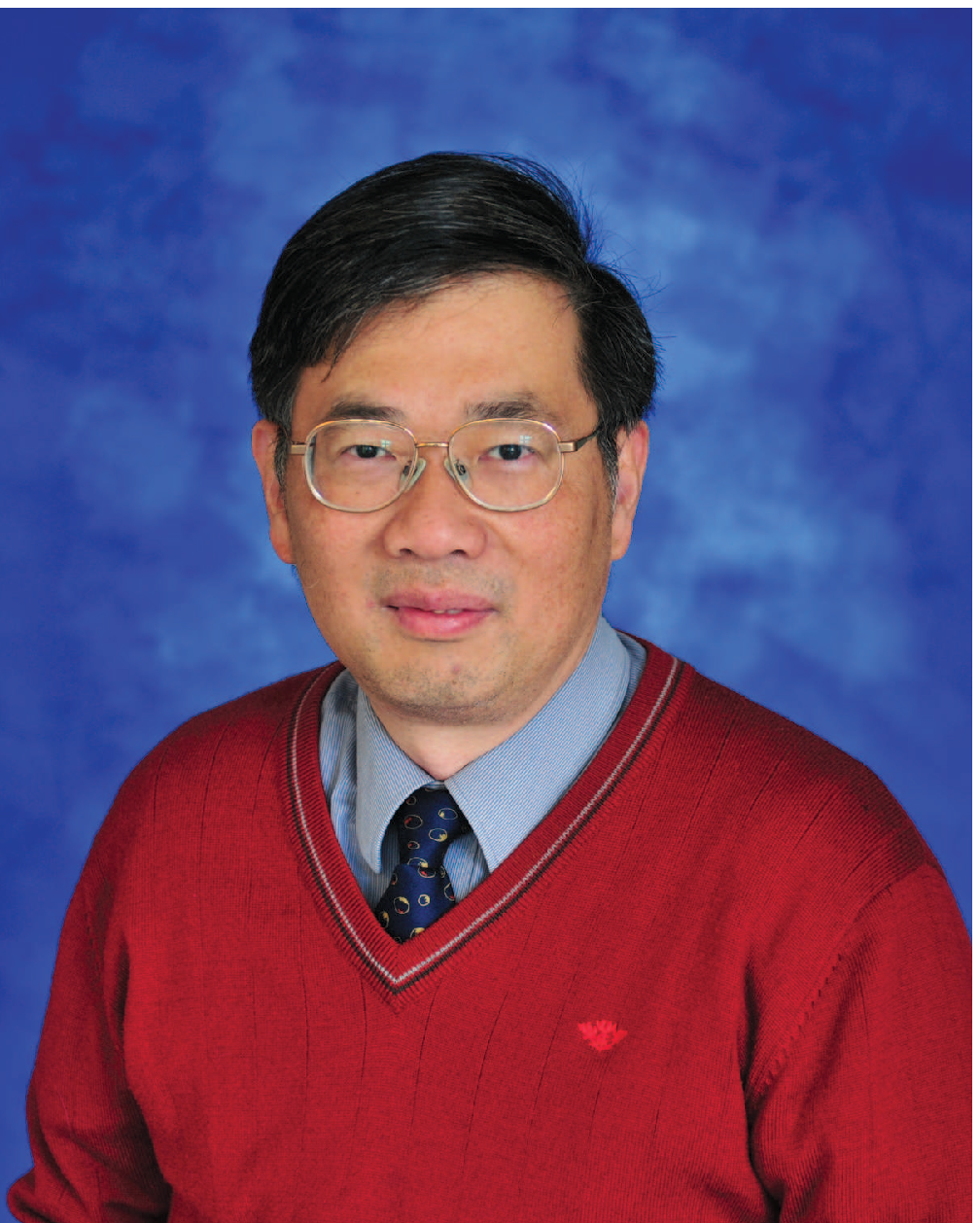}}]{Cheng-Shang Chang} (S'85-M'86-M'89-SM'93-F'04) received the B.S. degree from National Taiwan University, Taipei, Taiwan, in 1983, and the M.S. and Ph.D. degrees from Columbia University, New York, NY, USA, in 1986 and 1989, respectively, all in electrical engineering.

From 1989 to 1993, he was employed as a Research Staff Member with the IBM Thomas J. Watson Research Center, Yorktown Heights, NY, USA. Since 1993, he has been with the Department of Electrical Engineering, National Tsing Hua University, Taiwan, where he is a Tsing Hua Distinguished Chair Professor. He is the author of the book Performance Guarantees in Communication Networks (Springer, 2000) and the coauthor of the book Principles, Architectures and Mathematical Theory of High Performance Packet Switches (Ministry of Education, R.O.C., 2006). His current research interests are concerned with network science, big data analytics, mathematical modeling of the Internet, and high-speed switching.

Dr. Chang served as an Editor for Operations Research from 1992 to 1999, an Editor for the {\em IEEE/ACM TRANSACTIONS ON NETWORKING} from 2007 to 2009, and an Editor for the {\em IEEE TRANSACTIONS ON NETWORK SCIENCE AND ENGINEERING} from 2014 to 2017. He is currently serving as an Editor-at-Large for the {\em IEEE/ACM TRANSACTIONS ON NETWORKING}. He is a member of IFIP Working Group 7.3. He received an IBM Outstanding Innovation Award in 1992, an IBM Faculty Partnership Award in 2001, and Outstanding Research Awards from the National Science Council, Taiwan, in 1998, 2000, and 2002, respectively. He also received Outstanding Teaching Awards from both the College of EECS and the university itself in 2003. He was appointed as the first Y. Z. Hsu Scientific Chair Professor in 2002. He received the Merit NSC Research Fellow Award from the National Science Council, R.O.C. in 2011. He also received the Academic Award in 2011 and the National Chair Professorship in 2017 from the Ministry of Education, R.O.C. He is the recipient of the 2017 IEEE INFOCOM Achievement Award.
\end{IEEEbiography}




\end{document}